\documentclass[letterpaper, 10 pt, journal, twoside]{IEEEtran}
\usepackage{graphics} 
\usepackage{subfigure} 
\usepackage{epsfig} 
\usepackage{amsmath} 
\usepackage{amssymb}  
\usepackage{multirow}

\usepackage{makecell}

\usepackage{cite}
\usepackage{array}
\usepackage{url}

\usepackage{booktabs}
\usepackage{footnote}
\usepackage{siunitx}

\usepackage{etoolbox}
\makeatletter
\patchcmd{\@makecaption}
  {\scshape}
  {}
  {}
  {}
\makeatother

\makeatletter
\let\NAT@parse\undefined
\makeatother
\usepackage{hyperref}  

\IEEEoverridecommandlockouts
\IEEEpubid{\makebox[\columnwidth]{\copyright~2023 IEEE. Personal use of this material is permitted. \hfill} \hspace{\columnsep}\makebox[\columnwidth]{ }}

\begin{document}
\title{
RGBlimp:\! Robotic\! Gliding\! Blimp\! -\! Design,\! \\Modeling,\! Development,\! and\! Aerodynamics\! Analysis
}
%
%
%

\author{Hao Cheng, Zeyu Sha, Yongjian Zhu, and Feitian Zhang%
\thanks{Manuscript received April 23, 2023; accepted September 9, 2023. 
This paper was recommended for publication by Associate Editor T. Asfour and  Editor C. Gosselin upon evaluation of the reviewers' comments. 
This work was supported by the Fundamental Research Funds for the Central Universities, Peking University (7100604091).
\textit{(Corresponding author: Feitian Zhang.)}} 
\thanks{The authors are with the Department of Advanced Manufacturing and Robotics, College of Engineering, Peking University, Beijing 100871, China (e-mail: h-cheng@stu.pku.edu.cn; schahzy@stu.pku.edu.cn; yongjianzhu@pku.edu.cn; feitian@pku.edu.cn).}%
\thanks{Digital Object Identifier (DOI): 10.1109/LRA.2023.3318128}
}
%
%

\markboth{IEEE Robotics and Automation Letters. Preprint Version. Accepted September, 2023}
{Cheng \MakeLowercase{\textit{et al.}}: 
RGBlimp: Robotic Gliding Blimp - Design, Modeling, Development, and Aerodynamics Analysis} 

%


\maketitle

\begin{abstract}
A miniature robotic blimp, as one type of lighter-than-air aerial vehicle, has attracted increasing attention in the science and engineering field for its long flight duration and safe aerial locomotion. 
While a variety of miniature robotic blimps have been developed over the past decade, most of them utilize the buoyant lift and neglect the aerodynamic lift in their design, thus leading to a mediocre aerodynamic performance, particularly in terms of aerodynamic efficiency and aerodynamic stability. 
This letter proposes a new design of miniature robotic blimp that combines desirable features of both a robotic blimp and a fixed-wing glider, named the \textit{Robotic~Gliding~Blimp}, or RGBlimp. 
This robot, equipped with an envelope filled with helium and a pair of wings, uses an internal moving mass and a pair of propellers for its locomotion control. 
This letter presents the design, dynamic modeling, prototyping, and system identification of the RGBlimp. 
To the best of the authors' knowledge, this is the first effort to systematically design and develop such a miniature robotic blimp with moving mass control and hybrid lifts. 
Experimental results are presented to validate the design and the dynamic model of the RGBlimp. 
Analysis of the RGBlimp aerodynamics is conducted which confirms the performance improvement of the proposed RGBlimp in aerodynamic efficiency and flight stability. 
\end{abstract}

\begin{IEEEkeywords}
Aerial systems: mechanics and control, dynamics, mechanism design. 
\end{IEEEkeywords}

%
\IEEEpeerreviewmaketitle

\section{Introduction} 
\IEEEPARstart{M}{iniature} robotic blimps have attracted a rapidly growing interest in the science and engineering communities with their technological advances in actuation, sensing, and control \cite{GTMAB2}-\!\!\cite{blimpProjector}. 
The miniature robotic blimp, as one kind of lighter-than-air (LTA) aerial vehicle, has unique advantages such as long endurance in the air, enhanced safety in human-robot interaction, and low acoustic noise level in locomotive operation with great potential in applications including search and rescue \cite{darpa}, environmental monitoring \cite{blimpMonitoring}, and interactive entertainment \cite{blimpIROS13}. 

Whereas a miniature robotic blimp uses buoyant lift for long flight duration and locomotion safety, the aerodynamic performance of such a robot is quite limited, especially in low-speed cruising scenarios. 
A case in point is that the control surface such as the rudder, the elevator, and the aileron, while often used in the control of high-speed aerial vehicles such as large-scale airships and airplanes, has significantly decreased effectiveness in control of low-speed miniature robotic blimps. 
On the other hand, gliding mechanism has been applied to aerial vehicles to achieve high aerodynamic efficiency. For example, fixed-wing gliders \cite{fixedwingSurvey} and hybrid airships \cite{HA-aero}-\!\!\cite{HybridSurvey} use a pair of wings to provide aerodynamic lift for gliding, but generally require high-speed flight to generate sufficient lift. 

\begin{figure}[t]
      \centering
      \includegraphics[scale=0.205]{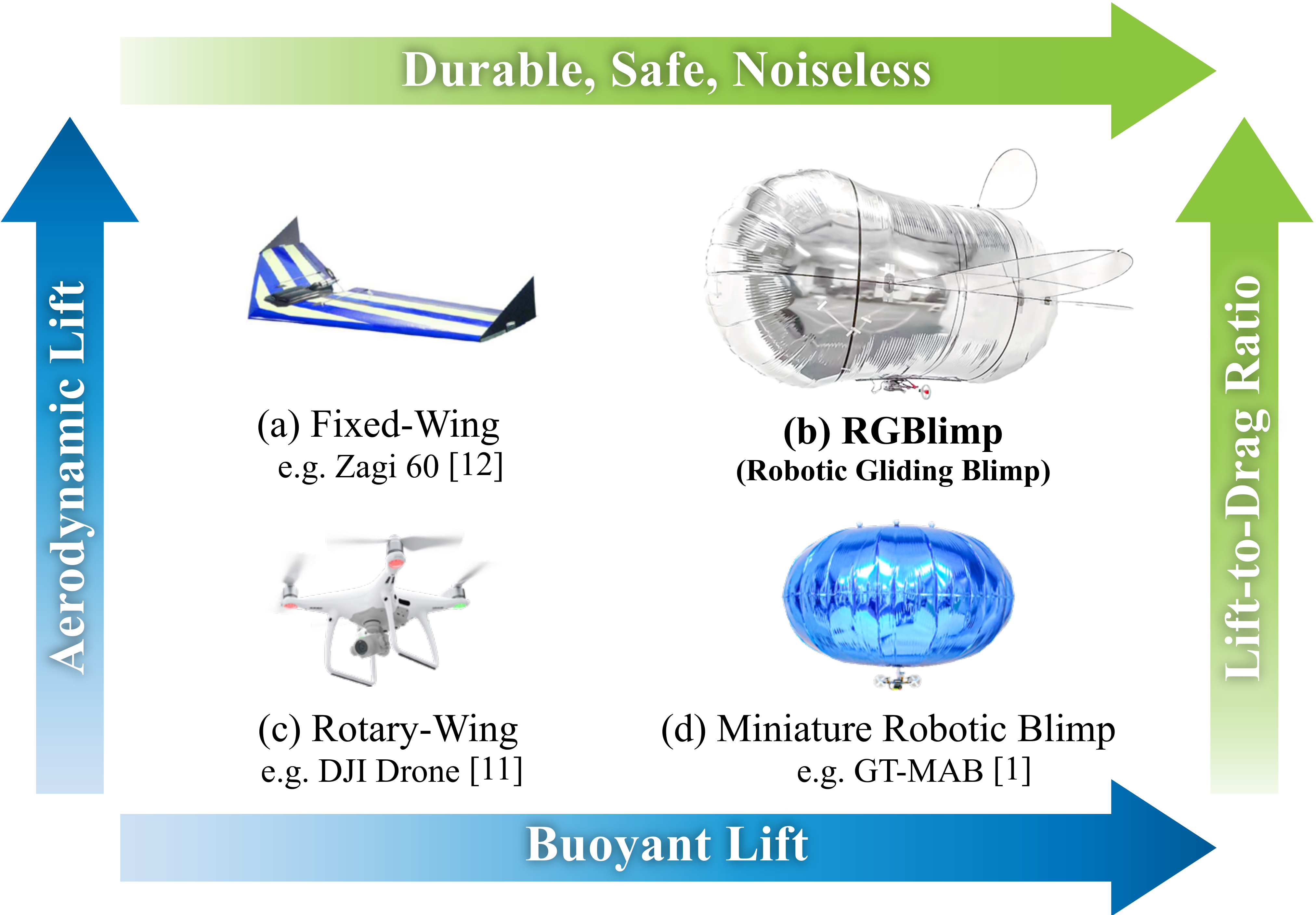}
      \vspace{-0.5mm} 
      \caption{The schematic of the conceptual design of the RGBlimp inspired from existing miniature aerial vehicles. While with advantages in buoyant lift, robotic blimps (d) are more durable, safer, and less noisy, their aerodynamic performance is rarely considered. In contrast, fixed-wing gliders (a) are designed to take advantage of the gliding mechanism with a higher lift-to-drag ratio but generally lose aerodynamic efficiency at low speed. Combining the advantages of buoyant and aerodynamic lift, the RGBlimp (b) as a hybrid aerial robot is designed to adapt to low-speed flight with improved aerodynamic performance. } 
      \label{fig.1}
      \vspace{-1.mm} 
\end{figure}

This letter proposes a new miniature robotic blimp, named Robotic Gliding Blimp (RGBlimp) that is equipped with a helium-filled envelope for buoyant lift and a pair of wings for aerodynamic lift, aiming to combine the desirable features of a conventional blimp and a fixed-wing glider. 
More importantly, the RGBlimp adopts a moving mass for attitude control, inspired by the centroid control of fly squirrels in gliding \cite{bio}, which is expected to have an improved flight control performance over control surfaces (elevators) at low-speed locomotion. 
To our best knowledge, the RGBlimp is the \textbf{first} miniature robotic blimp in the literature with moving mass control and hybrid buoyant-aerodynamic lift. 
Following the design and dynamic modeling, this letter proposes a method of system identification. 
The experimental results are presented to validate the proposed design and the derived dynamic model. 
Analysis is conducted upon the difference in the aerodynamic performance between the proposed RGBlimp and the conventional aerial robots such as the miniature robotic blimp and the fixed-wing aerial vehicle. 

This letter has three main contributions as follows. 
\begin{itemize}
    \setlength{\itemsep}{3pt}
    \item First, the design of the RGBlimp, a novel miniature robotic blimp, is proposed. 
          Particularly, the RGBlimp has a unique hybrid aerodynamic-buoyant lift design, and a moving mass control for low-speed flight. 
    \item Second, the dynamic model that differs from the existing work by considering the dynamic effects of the moving mass is derived using Euler's equations of motion. 
          System identification of the RGBlimp, particularly aerodynamic identification, is proposed and tested using nonlinear regression on steady flight data. 
    \item Third, an RGBlimp prototype is developed and experimentally tested to validate the proposed hybrid lift design and the dynamic modeling. 
          In addition, the experimental results confirm our design in improving the aerodynamic performances when compared with other conventional aerial robots. 
\end{itemize}


\section{Design and Dynamic Modeling}
\label{Sec:Model}

\subsection{Design of the RGBlimp}
An RGBlimp combines the desirable features of both a miniature blimp and a fixed-wing glider, designed for low-speed flight applications with long endurance, high safety, low noise, and enhanced aerodynamic performance. 
The RGBlimp stays aloft using the buoyant lift with a lightweight lifting gas (e.g., helium) and the aerodynamic lift with a pair of fixed wings. 
While vertical thrusters provide altitude control in most existing miniature robotic blimps, the RGBlimp does not include those thrusters but rather relies on gliding with pitch regulation for altitude control aiming for a higher lift-to-drag ratio (L/D) and higher aerodynamic efficiency. 

The RGBlimp has two actuation systems for propulsion and orientation control. 
Specifically, a pair of small-sized propellers provide propulsion and yaw adjustment. 
The moving mass controls the attitude by adjusting the mass distribution of the robot. 


\subsection{Dynamic Modeling}
This section derives the dynamic model of the RGBlimp. 
The non-zero displacement of the center of gravity (CG) from the center of buoyancy (CB) and the motion of the moving mass will be considered, which is often ignored in the dynamic modeling of most existing miniature robotic blimps \cite{GTMAB2},\!\cite{blimpICRA19}. 

\begin{figure}[thpb]
      \centering
      \vspace{-0mm}
      \subfigbottomskip=2pt
      \subfigure[The Mass Distribution.]{
      \label{fig.2a}
      \includegraphics[scale=0.245]{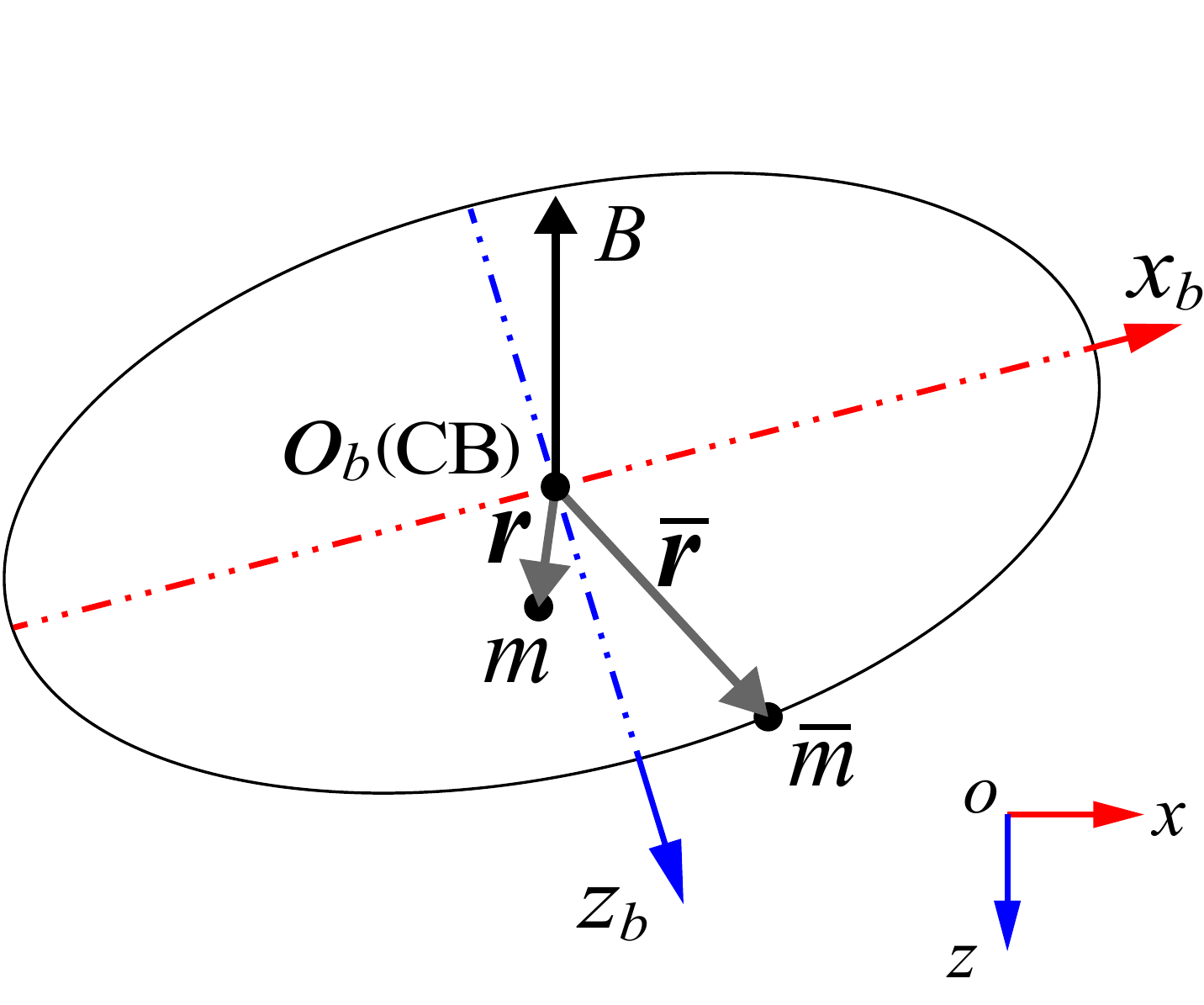} 
      } 
      \subfigure[The Reference Frames.]{
      \label{fig.2b}
      \includegraphics[scale=0.245]{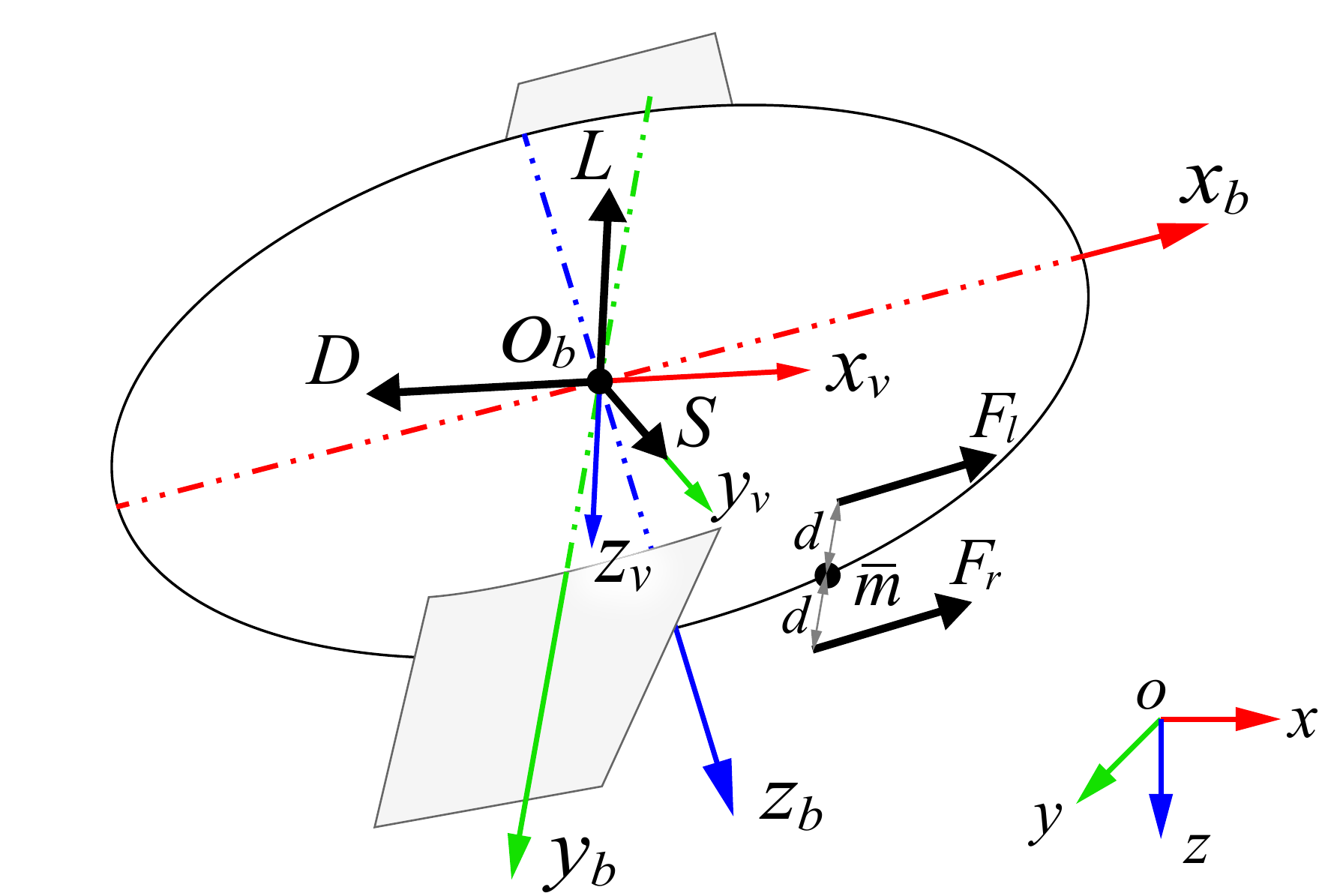}
      }
      \vspace{0.5mm}
      \caption{Illustration of the RGBlimp model. (a) The mass distribution (side~view), including the stationary mass $m$ and the moving mass $\bar{m}$. (b) The coordinate frames, differential propulsion, and aerodynamic forces. }
      \label{fig.2}
\end{figure}

We model the RGBlimp as a rigid-body dynamical system of six degrees of freedom with external forces and moments exerted by the moving mass, gravity, and surrounding passing air. 
Figure~\ref{fig.2a} shows the mass distribution of the robot. 
The moving mass is denoted as $\bar{m}$ with a controllable displacement $\bar{\boldsymbol{r}}$ with respect to CB; the mass of the stationary body (excluding the moving mass $\bar{m}$) is denoted as $m$ with a constant displacement $\boldsymbol{r}$ with respect to CB. 

\vspace{2mm}

Figure\!~\ref{fig.2b} shows the reference frames as well as the external forces. 
Following the literature \cite{flightDynamics}, the pose of the robot including the position and the orientation is defined and represented with respect to the earth-fixed inertial reference frame that is denoted by $O\text{-}xyz$. 
The translational and rotational velocities of the RGBlimp are defined and represented with respect to the body-fixed frame that is denoted by $O_b\text{-}x_by_bz_b$, whose origin $O_b$ is defined at CB. 
The thrusts $F_l$ and $F_r$ are assumed to be parallel to the $x_b$-axis. 
 
The orientation of the RGBlimp is described by the Euler angles including the roll angle $\phi$, the pitch angle $\theta$, and the yaw angle $\psi$. 
Let $\boldsymbol{e}_i=[\phi,\theta,\psi]^\mathrm{T}$ stand for the orientation. 
Let $\boldsymbol{p}_i=[x,y,z]^\mathrm{T}$ stand for the position vector, describing the displacement of the origin of the inertial frame $O$ relative to the origin of the body-fixed frame $O_b$. 
Let $\boldsymbol{v}_b=[u,v,w]^\mathrm{T}$ and $\boldsymbol{\omega}_b=[p,q,r]^\mathrm{T}$ stand for the translational and rotational velocity vectors of the body-fixed frame relative to the inertial frame, respectively. 
The kinematics of the RGBlimp is described by
\begin{align}
    \displaystyle
    \dot{\boldsymbol{p}}_i=\boldsymbol{R} \boldsymbol{v}_b,\hspace{5mm}
    \dot{\boldsymbol{e}}_i=\boldsymbol{J} \boldsymbol{\omega}_b.\label{eq.1}
\end{align}
Here, $\boldsymbol{R}=\boldsymbol{R}_\psi\boldsymbol{R}_\theta\boldsymbol{R}_\phi$ represents the rotation operation from the body-fixed frame to the inertial frame using the roll-pitch-yaw (RPY) sequence. 
The angular velocity $\boldsymbol{\omega}_b$ and the Euler angle changing rate $\dot{\boldsymbol{e}}_i$ are related via the transformation matrix $\boldsymbol{J}$ \cite{book1993}, i.e., 
\begin{equation}
    \boldsymbol{J}=\left[\begin{array}{ccc}
        1 & \sin \phi \tan \theta & \cos \phi \tan \theta \\
        0 & \cos \phi & -\sin \phi \\
        0 & \sin \phi / \cos \theta & \cos \phi / \cos \theta
\end{array}\right]\!.
	\label{eq.2}
\end{equation}

Based on Euler's first and second laws of motion, the dynamics of the RGBlimp are derived as follows
\begin{align}
    \displaystyle
    \boldsymbol{F}_b &= \hspace{0.5mm}m\left( \dot{\boldsymbol{v}}_b + \boldsymbol{\omega}_b\!\times\!\boldsymbol{v}_b + \boldsymbol{a} \right),\label{eq.4}\\
    \boldsymbol{T}_b &= \boldsymbol{I}\dot{\boldsymbol{\omega}}_b + \boldsymbol{\omega}_b\!\times\!\left( \boldsymbol{I}\boldsymbol{\omega}_b \right) + m\boldsymbol{r}\!\times\!\left( \dot{\boldsymbol{v}}_b + \boldsymbol{\omega}_b\!\times\!\boldsymbol{v}_b \right).\label{eq.5}
\end{align}
Here, $m$ is the stationary body mass; $\boldsymbol{I}$ is the stationary body inertia tensor about the origin of the body-fixed frame $O_b$. 
$\boldsymbol{F}_b$ and $\boldsymbol{T}_b$ are the total external force vector and moment vector acting on the stationary mass, respectively, expressed in the body-fixed frame. $\boldsymbol{a}$ is the sum of Euler acceleration and centripetal acceleration of the stationary body mass $m$, expressed in the body-fixed frame \cite{modelBlimpICRA98}\cite{mechanics}, i.e., 
\begin{equation}
    \boldsymbol{a}= \dot{\boldsymbol{\omega}}_b\!\times\!\boldsymbol{r}+\boldsymbol{\omega}_b\!\times\!(\boldsymbol{\omega}_b\!\times\!\boldsymbol{r}) .
	\label{eq.6}
\end{equation}
The total external force is calculated as
\begin{equation}
    \boldsymbol{F}_b = (F_l\!+\!F_r)\boldsymbol{i}_b + (mg-B)\boldsymbol{R}^\mathrm{T}\boldsymbol{k} + \boldsymbol{F}_\mathrm{aero}+\boldsymbol{f} .
	\label{eq.7}
\end{equation}
Here, $F_l$ and $F_r$ represent the thrust forces generated by the left and right propellers, respectively. 
$\boldsymbol{i}_b\!=\![1,0,0]^\mathrm{T}$ and $\boldsymbol{k}\!=\![0,0,1]^\mathrm{T}$ are the unit vectors along the $O_bx_b$ and $Oz$ axes, respectively. 
$B\!=\!\rho gV_\mathrm{He}$ stands for the buoyant force, where $\rho$ is the density of the air and $V_\mathrm{He}$ is the volume of the helium gas. 
$\boldsymbol{F}_\mathrm{aero}$ represents the aerodynamic forces acting on the robot expressed in the body-fixed frame, which will be discussed in detail in Sec.\!~\ref{subsec:AeroModel}.

By D'Alembert's principle, the term $\boldsymbol{f}$ in \eqref{eq.7} is the total force acting on the stationary mass, exerted by the moving mass expressed in the body-fixed frame, i.e.,
\begin{equation}
    \boldsymbol{f} = \bar{m}g\boldsymbol{R}^\mathrm{T}\boldsymbol{k} - \bar{m}\dot{\bar{\boldsymbol{v}}}, 
	\label{eq.8}
\end{equation}
where $\dot{\bar{\boldsymbol{v}}}$ is the time derivative of the velocity of the moving mass $\bar{m}$, expressed in the body-fixed frame, i.e.,
\begin{equation}
    \begin{aligned}
	   \displaystyle
       \dot{\bar{\boldsymbol{v}}} =&\ \boldsymbol{R}^\mathrm{T}\frac{\mathrm{d}}{\mathrm{d} t}\boldsymbol{R}\left(\boldsymbol{v}_b + \dot{\bar{\boldsymbol{r}}} + \boldsymbol{\omega}_b\!\times\!\bar{\boldsymbol{r}}\right) \\
                              =&\ \dot{\boldsymbol{v}}_b + \boldsymbol{\omega}_b\!\times\!\boldsymbol{v}_b + \ddot{\bar{\boldsymbol{r}}}+2\boldsymbol{\omega}_b\!\times\!\dot{\bar{\boldsymbol{r}}}\\
                              &+ \dot{\boldsymbol{\omega}}_b\!\times\!\bar{\boldsymbol{r}} + \boldsymbol{\omega}_b \!\times\! \left( \boldsymbol{\omega}_b\!\times\!\bar{\boldsymbol{r}}\right).
    \end{aligned}
	\label{eq.9}
\end{equation}

The total external moment $\boldsymbol{T}_b$ is calculated as
\begin{equation}
    \begin{aligned}
	   \displaystyle
       \boldsymbol{T}_b =&\ (F_l\!+\!\!F_r)\bar{r}_z\boldsymbol{j}_b + (F_l\!-\!\!F_r)d\,\boldsymbol{k}_b\\
                         &+\boldsymbol{r}\!\times\! mg\boldsymbol{R}^\mathrm{T}\boldsymbol{k} + \boldsymbol{T}_\mathrm{aero} + \bar{\boldsymbol{r}}\!\times\!\boldsymbol{f} .
    \end{aligned}
	\label{eq.10}
\end{equation}
Here, $\boldsymbol{j}_b\!=\![0,1,0]^\mathrm{T}$ and $\boldsymbol{k}_b\!=\![0,0,1]^\mathrm{T}$ are the unit vectors along the $O_by_b$ and $O_bz_b$ axis, respectively. 
$\bar{r}_z$ is the third element of the vector $\bar{\boldsymbol{r}}$. 
$d$ is the distance from each propeller to the $O_bx_b$ axis (Fig.\!~\ref{fig.2b}). 
$\boldsymbol{T}_\mathrm{aero}$ is the aerodynamic moment acting on the robot expressed in the body-fixed frame, which will be discussed in detail in Sec.\!~\ref{subsec:AeroModel}. 

Combining \eqref{eq.1}-\eqref{eq.10}, we rewrite the full dynamic model of the RGBlimp in terms of the system states $\boldsymbol{p}_i, \boldsymbol{e}_i, \boldsymbol{v}_b, \boldsymbol{\omega}_b, \bar{\boldsymbol{r}}, \dot{\bar{\boldsymbol{r}}}$ and the control inputs $F_l, F_r, \bar{F}_x, \bar{F}_y, \bar{F}_z$ in a vector form as follows
\begin{align}
    \begin{bmatrix}
      \dot{\boldsymbol{p}}\\
      \dot{\boldsymbol{e}}\\
      \dot{\bar{\boldsymbol{r}}}
    \end{bmatrix}\hspace{1mm}
    &=\hspace{3.2mm}
    \begin{bmatrix}
      \boldsymbol{R} \boldsymbol{v}_b\\
      \boldsymbol{J} \boldsymbol{\omega}_b\\
      \dot{\bar{\boldsymbol{r}}}
    \end{bmatrix}, \label{eq.11}\\[+1mm]
    \begin{bmatrix}
      \dot{\boldsymbol{v}}_b\\
      \dot{\boldsymbol{\omega}}_b\\
      \ddot{\bar{\boldsymbol{r}}}
    \end{bmatrix}
    &=
    \boldsymbol{A}\!
    \begin{bmatrix}
      \,\tilde{\boldsymbol{f}}\ \\
      \,\tilde{\boldsymbol{t}}\ \\
      \,\boldsymbol{0}_{3\!\times\!1}\ 
    \end{bmatrix}+
    \boldsymbol{B}\!
    \begin{bmatrix}
      F_l\\
      F_r\\
      \bar{\boldsymbol{F}}
    \end{bmatrix}, \label{eq.12}
\end{align}
\vspace{-2mm}
\noindent where
\begin{align}
    \hspace{1mm}\tilde{\boldsymbol{f}}\hspace{0.5mm} &= (m\!+\!\bar{m})\boldsymbol{v}_b\!\times\!\boldsymbol{\omega}_b+
    (\boldsymbol{\omega}_b\!\times\!\boldsymbol{l}_g)\!\times\!\boldsymbol{\omega}_b\hspace{0.8mm}+\nonumber\\
    &\hspace{4.5mm}(mg\!+\!\bar{m}g\!-\!B)\boldsymbol{R}^\mathrm{T}\boldsymbol{k} + \boldsymbol{F}_\mathrm{aero} + 2\bar{m}\dot{\bar{\boldsymbol{r}}}\!\times\!\boldsymbol{\omega}_b, \label{eq.13}\\[+1.5mm]
    \hspace{1mm}\tilde{\boldsymbol{t}}\hspace{0.7mm} &= 
    \boldsymbol{l}_g\!\times\!(\boldsymbol{v}_b\!\times\!\boldsymbol{\omega}_b) + 
    \left(\boldsymbol{I}\!-\!\bar{m} (\boldsymbol{\bar{r} }^\times)^2\right)\boldsymbol{\omega}_b\!\times\!\boldsymbol{\omega}_b + \nonumber\\
    &\hspace{5mm}\boldsymbol{l}_g\!\times\!g\boldsymbol{R}^\mathrm{T}\boldsymbol{k} + \boldsymbol{T}_\mathrm{aero} + 2\bar{m}\bar{\boldsymbol{r}}\!\times\!(\dot{\bar{\boldsymbol{r}}}\!\times\!\boldsymbol{\omega}_b), \label{eq.14}
\end{align}
\vspace{-6mm}
\begin{align}
    \boldsymbol{A}
    &=\!
    \hspace{3.5mm}
    \begin{bmatrix}
      (m\!+\!\bar{m})\boldsymbol{1}_{3}&  -\boldsymbol{l}_g^\times & \bar{m} \boldsymbol{1}_{3}\\[+0.5mm]
      \boldsymbol{l}_g^\times&  \boldsymbol{I}\!-\!\bar{m} (\boldsymbol{\bar{r} }^\times)^2&\,\bar{m} \boldsymbol{\bar{r} }^\times\\[+0.5mm]
      \boldsymbol{0}_{3}&  \boldsymbol{0}_{3}& \boldsymbol{1}_{3}
    \end{bmatrix}^{\!-\!1}\!\!\!, \label{eq.15}\\[+1mm]
    \boldsymbol{B}
    &=\!
    \boldsymbol{A}
    \begin{bmatrix}
       1 & 0 & 0 & 0 & \bar{r}_z & \bar{r}_y\!+\!d &\multirow{2}{*}{$\boldsymbol{0}_{2\!\times\!3}$}\\
       1 & 0 & 0 & 0 & \bar{r}_z & \bar{r}_y\!-\!d &\\
       \multicolumn{6}{c}{\ \boldsymbol{0}_{3\!\times\!6}} & \boldsymbol{1}_{3}
    \end{bmatrix}^\mathrm{T}\!\!. \label{eq.16}
\end{align}
Here, $\bar{\boldsymbol{F}} = [\bar{F}_x,\bar{F}_y,\bar{F}_z]^\mathrm{T}\!=\ddot{\bar{\boldsymbol{r}}}$ is the relative acceleration of  the moving mass $\bar{m}$, considered as part of the control inputs. 
$\boldsymbol{l}_g\!=\!m\boldsymbol{r}\!+\!\bar{m}\bar{\boldsymbol{r}}$ satisfies $\boldsymbol{r}\!_g\!=\!\boldsymbol{l}_g/(m\!+\!\bar{m})$ where $\boldsymbol{r}\!_g$ is the CG of the robot. 
$\boldsymbol{0}$ is the zero vector or matrix. $\boldsymbol{1}$ is the identity matrix. The determinant of $\boldsymbol{A}$ is always greater than zero, hence $\boldsymbol{A}\!^{-1}$ is always well-defined. 

\subsection{Aerodynamic Model}
\label{subsec:AeroModel}
To model the aerodynamics, we define the velocity reference frame $O_b\text{-}x_vy_vz_v$. 
Specifically, the $O_bx_v$ axis is along the direction of the velocity, and the $O_bz_v$ is perpendicular to $O_bx_v$ in the plane of symmetry of the robot, as shown in Fig.\!~\ref{fig.2b}. 
Rotation matrix $\boldsymbol{R}_v^b$ represents the rotation operation from the velocity reference frame to the body-fixed frame, i.e., 
\begin{equation}
    \boldsymbol{R}_v^b=\left[\begin{array}{ccc}
        \cos\alpha \cos\beta & -\cos\alpha \sin\beta & -\sin\alpha \\
        \sin\beta & \cos\beta & 0 \\
        \sin\alpha\cos\beta & -\sin\alpha \sin\beta & \cos\alpha
\end{array}\right],
	\label{eq.17}
\end{equation}
where $\alpha\!=\!\arctan(w/u)$ is the angle of attack, $\beta\!=\!\arcsin(v/V)$ is the sideslip angle, and $V$ is the velocity magnitude.

Expressed in the velocity reference frame, the aerodynamic forces include the drag force $D$, the side force $S$, and the lift force $L$\hspace{0.5mm}, as shown in Fig.\!~\ref{fig.2b}; the aerodynamic moments include the rolling moment $M_1$, the pitching moment $M_2$, and the yaw moment $M_3$. Then we have
\begin{align}
    \displaystyle
    \boldsymbol{F}_\mathrm{aero}&=\boldsymbol{R}_v^b[-D\hspace{0.3mm},\hspace{1.8mm}\ S\hspace{0.8mm},\hspace{0.2mm} -L\hspace{0.9mm}]^\mathrm{T},\label{eq.18}\\
    \boldsymbol{T}_\mathrm{aero}&=\boldsymbol{R}_v^b[\hspace{1.1mm}M_1,\ M_2,\ M_3]^\mathrm{T}.\label{eq.19}
\end{align}

The aerodynamic forces and moments are dependent on the angle of attack $\alpha$, the sideslip angle $\beta$, the velocity magnitude $V$ \cite{dynamic}\!\cite{aircraft}, i.e.,
\begin{align}
	\hspace{0.3mm}D\hspace{-0.3mm}\ &= 1/2\, \rho V^2 \!A\ C_D(\alpha,\!\beta),\label{eq.20}\\ 
	S\ &= 1/2\, \rho V^2 \!A\ C_S(\alpha,\!\beta),\\
	L\ &= 1/2\, \rho V^2 \!A\ C_L(\alpha,\!\beta),\\
	M_1 &= 1/2\, \rho V^2 \!A\ C_{M_1}(\alpha,\!\beta) + K_1p,\\
	M_2 &= 1/2\, \rho V^2 \!A\ C_{M_2}(\alpha,\!\beta) + K_2q,\\
	M_3 &= 1/2\, \rho V^2 \!A\ C_{M_3}(\alpha,\!\beta) + K_3r.\label{eq.25}
\end{align}
Here, scalar $A$ is the reference area, which characterizes the design characteristic of the robot. 
$K_1$, $K_2$, $K_3 \in \mathbb{R}^-$ are the rotational damping coefficients. 
The terms in the form of $C_\Box(\alpha,\!\beta)$ are the corresponding aerodynamic coefficients.

\subsection{Steady-State Equations} 
Given fixed control inputs, we obtain steady-state equations that describe the steady flight motion of the RGBlimp. 
Assuming the velocity of moving mass $\dot{\bar{\boldsymbol{r}}}$ is zero, the steady-state equations are obtained by setting the time derivatives of \eqref{eq.12} to zero, yielding
\begin{align}
    \displaystyle
    \!0=&\,\boldsymbol{F}_{aero} + (m\!+\!\bar{m})\boldsymbol{v}_b\!\times\!\boldsymbol{\omega}_b+
    (\boldsymbol{\omega}_b\!\times\!\boldsymbol{l}_g)\!\times\!\boldsymbol{\omega}_b\hspace{0.8mm}+\nonumber\\
    &(mg\!+\!\bar{m}g\!-\!B)\boldsymbol{R}^\mathrm{T}\boldsymbol{k} + (F_l\!+\!F_r)\boldsymbol{i}_b,\label{eq.26} \\[+1mm]
    \!0=&\,\boldsymbol{T}_{aero} +\boldsymbol{l}_g\!\times\!(\boldsymbol{v}_b\!\times\!\boldsymbol{\omega}_b) + 
    \left(\boldsymbol{I}\!-\!\bar{m} (\boldsymbol{\bar{r} }^\times)^2\right)\boldsymbol{\omega}_b\!\times\!\boldsymbol{\omega}_b + \nonumber\\
    &\,\boldsymbol{l}_g\!\times\!g\boldsymbol{R}^\mathrm{T}\boldsymbol{k} + (F_l\!+\!F_r)[0,\bar{r}_z,\bar{r}_y]^\mathrm{T}\! + (F_l\!-\!F_r)d\boldsymbol{k}_b.\label{eq.27}
\end{align}

The steady flight when $F_l - F_r$ is nonzero corresponds to a spiraling motion, similar to the steady spiraling of underwater gliders \cite{glidingfish}. 
There are six independent states for describing the steady spiral motion, including $\theta,\phi,\dot{\psi},V,\alpha,\beta$, which could be uniquely determined by solving \eqref{eq.26}-\eqref{eq.27}. 
The steady spiraling flight changes into a planar straight-line flight when $F_l\!-\!F_r$, $\dot{\psi}$, and $\beta$ are all zeros.

\section{The RGBlimp Prototype and \\System Identification}
\label{sec:Aero}

\subsection{RGBlimp Prototype and Experimental Setup} 
\begin{figure}[thpb]
      \centering
      \vspace{-0mm}
      \includegraphics[scale=0.15]{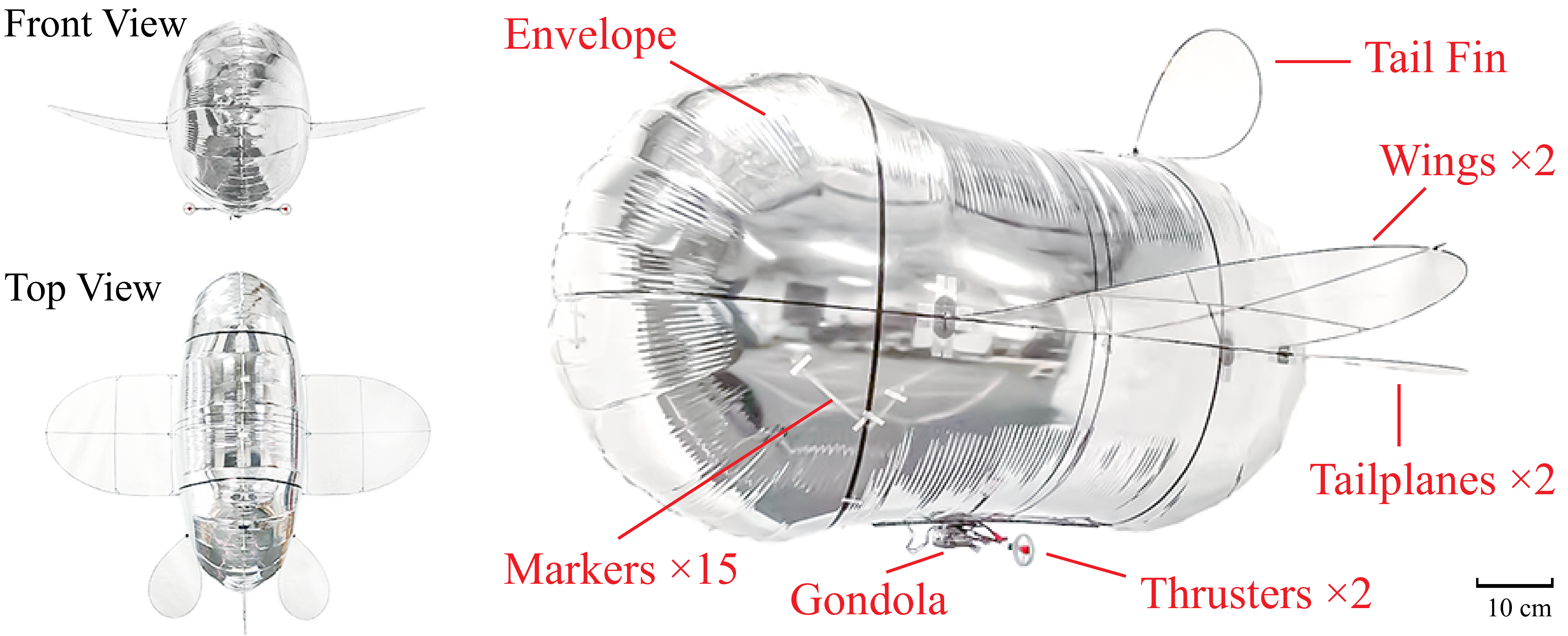}
      \vspace{-0mm}
      \caption{The RGBlimp prototype design includes an envelope, a pair of main wings, a tail fin, two tailplanes, and a gondola that consists of a pair of propellers, a controller unit, and a battery. }
      \label{fig.3}
\end{figure}

We successfully developed a prototype of the RGBlimp shown in Fig.\!~\ref{fig.3}. 
The net weight of the robot is $\SI{6.85}{g}$ and the dimension is $1.0\!\times\!1.1\!\times\!\SI{0.5}{m}$, designed to accommodate necessary hardware components and other required payload. 

The mass distribution of the RGBlimp prototype is detailed in Table~\ref{tab:mass}. 
The stationary mass $m$ mainly includes an envelope, a pair of main wings, a tail fin, and two tailplanes. 
Specifically, the envelope is a Mylar balloon filled with $\SI{125}{}$ liters of helium, providing roughly $\SI{152.04}{gf}$ buoyancy.
The pair of main wings takes an aerodynamic shape with $\SI{400}{mm}$ wingspan and $\SI{15}{\deg}$ dihedral angle, providing necessary aerodynamic lift and rolling stability. 
The vertical tail fin and horizontal tailplanes are mounted onto the tail section of the prototype. 
Active markers ($\SI{850}{nm}$ infrared LEDs) are used for the motion capture experiments. 

A battery, a control unit, and a pair of propellers are integrated into the gondola serving as the moving mass $\bar{m}$ that uses a linear actuator to move back and forth along a 3D-printed guide rail mounted onto the bottom of the RGBlimp prototype. 
The position of the moving mass is represented by $\bar{\boldsymbol{r}}\!=\!\bar{\boldsymbol{r}}_0\!+\!\Delta\bar{r}_x\boldsymbol{i}_b$ where $\bar{\boldsymbol{r}}_0$ is the designated initial position vector of the moving mass, $\Delta\bar{r}_x$ is the displacement along the longitudinal axis of the robot $x_b$, and $\boldsymbol{i}_b$ is the unit vector in the direction of $x_b$. 
The pair of propellers, mounted on the left and right sides of the gondola, provide differential propulsion for forward motion and yaw adjustment, with a distance of $d\!=\!\SI{150}{mm}$ from the $x\text{-}O\text{-}z$ plane. 

The propellers are powered by brushless DC motors \textit{SE0802-KV19000} and electronic speed controllers (ESC). 
The control unit that comprises a low-cost microcontroller \textit{ESP8266} utilizes the \textit{rosserial\_arduino} library and builds a \textit{Robot Operating System} (ROS) node for Wi-Fi communication with a nearby computer workstation. 

\begin{table}[t]
    \vspace{2mm}
    \caption{\vspace{0mm}Mass distribution of the RGBlimp prototype.}
    \label{tab:mass}
    \vspace{-2mm}
    \begin{center}
        \begin{tabular}{p{5.3cm}p{1.7cm}}
            \toprule
            Component & \hspace{2.4mm}Weight\\
            \midrule
            \textbf{Buoyancy}$^\text{a}$ $-B$ (Helium)                  &\hspace{1.72mm}$\textbf{-152.04\,gf}$\\[1.3mm]
            \textbf{Stationary Mass} $m$                                &\hspace{2.65mm}$\textbf{104.81\,g}$\\[0.3mm]
            \hspace{3mm}Helium                                          &\hspace{3.73mm}$\SI{21.19}{g}$\\
            \hspace{3mm}Envelop                                         &\hspace{3.73mm}$\SI{40.73}{g}$\\
            \hspace{3mm}Main Wings                                      &\hspace{5.21mm}$\SI{9.42}{g}\times 2$\\
            \hspace{3mm}Tail Wings (Tail Fin \& Tailplanes)             &\hspace{5.21mm}$\SI{2.29}{g}\times 3$\\
            \hspace{3mm}Guide Rail (for Moving Mass)                    &\hspace{3.73mm}$\SI{15.13}{g}$\\ 
            \hspace{3mm}Motion Capture Markers (Infrared LEDs)          &\hspace{5.21mm}$\SI{2.05}{g}$\\[1.3mm]
            \textbf{Moving mass} $\bar{m}$ (Gondola)        &\hspace{3.96mm}$\textbf{54.08\,g}$\\[0.3mm]
            \hspace{3mm}Battery                                         &\hspace{3.73mm}$\SI{16.80}{g}$\\
            \hspace{3mm}Control Unit \& Thrusters                       &\hspace{3.73mm}$\SI{32.31}{g}$\\
            \hspace{3mm}Additional Weight                               &\hspace{5.21mm}$\SI{4.97}{g}$\\[0.3mm]
            \midrule
            \textbf{Total} (Net$^\text{b}$)                             &\hspace{5.41mm}$\textbf{6.85\,g}$\\
            \bottomrule
        \end{tabular}
    \end{center}
    \vspace{-1.0mm}
    \scriptsize{$^\text{a}$The total buoyancy of helium in gram force ($\SI{}{gf}$).} 
    \scriptsize{$^\text{b}$The total net mass refers to the total weight remaining after deducting buoyancy, i.e., $m\!+\!\bar{m}\!-\!B$.}
\end{table}

\subsection{System Identification}
\label{subsec:AeroID}
First, we determine the system parameters excluding the aerodynamic coefficients. 
The gravitational constant $g$ is $\SI{9.80}{m/s^2}$. 
The air density $\rho$ is $\SI{1.219}{Kg/m^3}$, and the Reynolds number $Re$ is $\num{3.4e4}$.
We calculate the CG $\boldsymbol{r}$ by hanging the stationary part of the robot (without gondola) at selected points, then taking the intersection of these hanging strings, resulting in $\boldsymbol{r}\!=\!(-43.2,0.3,7.9)\SI{}{mm}$. 
The moving mass (gondola) $\bar{m}$ is treated as a point mass. 
Its initial position $\bar{\boldsymbol{r}}_0$ is gauged by the motion capture system $\bar{\boldsymbol{r}}_0\!=\!(74.7,0.6,238.0)\SI{}{mm}$. 
We model the mass distribution of the robot using computer aided design (CAD) and calculate the inertia tensor $\boldsymbol{I}\!=\!\mathrm{diag}(0.030,0.015,0.010)\SI{}{kg\!\cdot\!m^2}$. 
The mapping from the ESC throttle to thrust forces is calibrated experimentally. 

With the developed prototype, extensive flight experiments were conducted to identify the aerodynamic coefficients. 
Specifically, the robot was flown in a motion capture arena ($\text{L\hspace{0.3mm}}5.0\!\times\!\text{W\hspace{0.3mm}}4.0\!\times\!\text{H\hspace{0.3mm}}\SI{2.5}{m}$) with ten \textit{OptiTrack} cameras. 
The system captures data at $\SI{60}{Hz}$, with an accuracy of $\SI{0.76}{mm}$ RMS in position estimation error. 
A router facilitates real-time communication between the robot and a PC through ROS. 
In each trial, an electromagnetic releaser ensures consistency in the starting pose. 

The aerodynamic model of the RGBlimp is different from conventional miniature robotic blimps \cite{aeroID1} and fixed-wing aircraft \cite{aeroID2}. 
We conducted aerodynamic model identification using a series of steady flight tests with different displacements of the moving mass $\Delta\bar{r}_x$. 
 
\begin{figure}[thpb]
      \centering
      \vspace{-1mm}
      \subfigbottomskip=2pt
      \subfigure[The Straight Flight.]{
      \label{fig.4a}
      \includegraphics[scale=0.43]{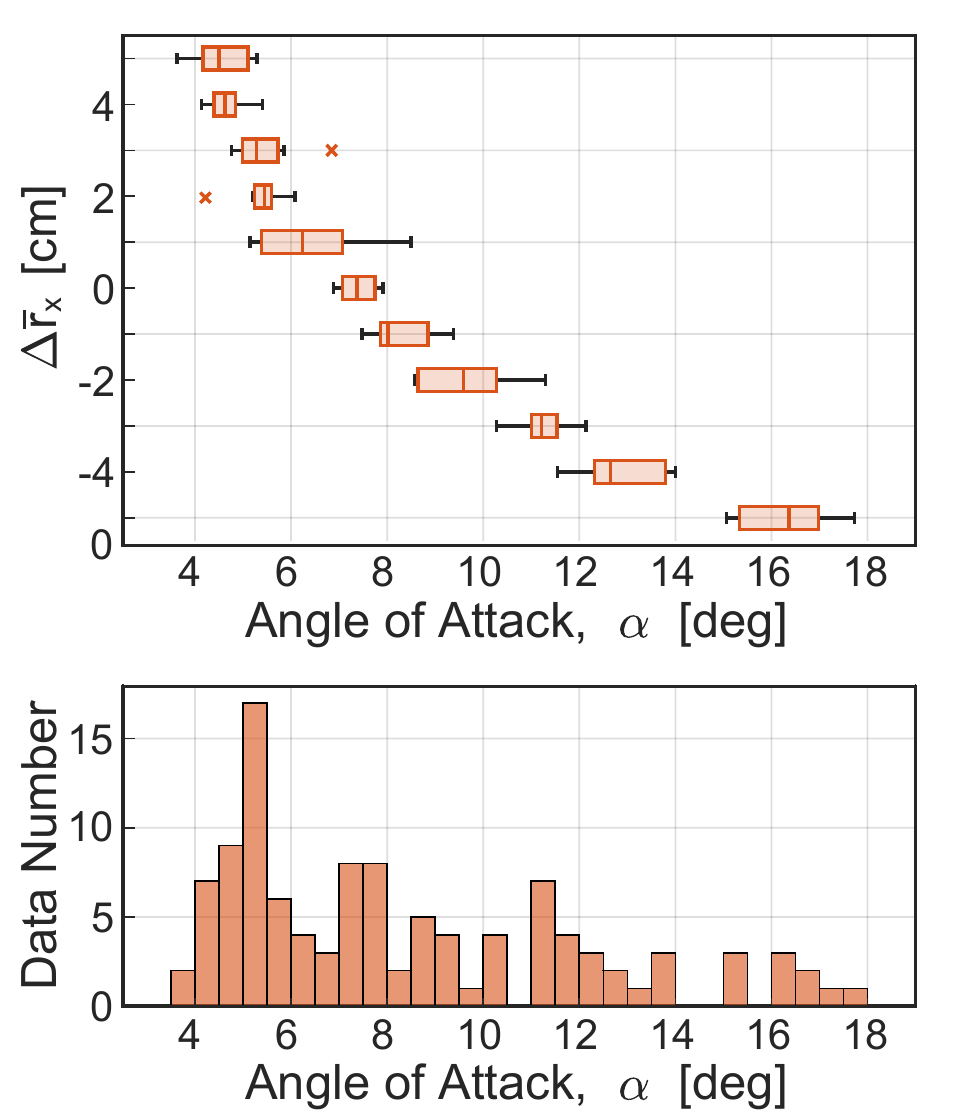} 
      } \hspace{-5mm}
      \subfigure[The Spiral Flight.]{
      \label{fig.4b}
      \includegraphics[scale=0.43]{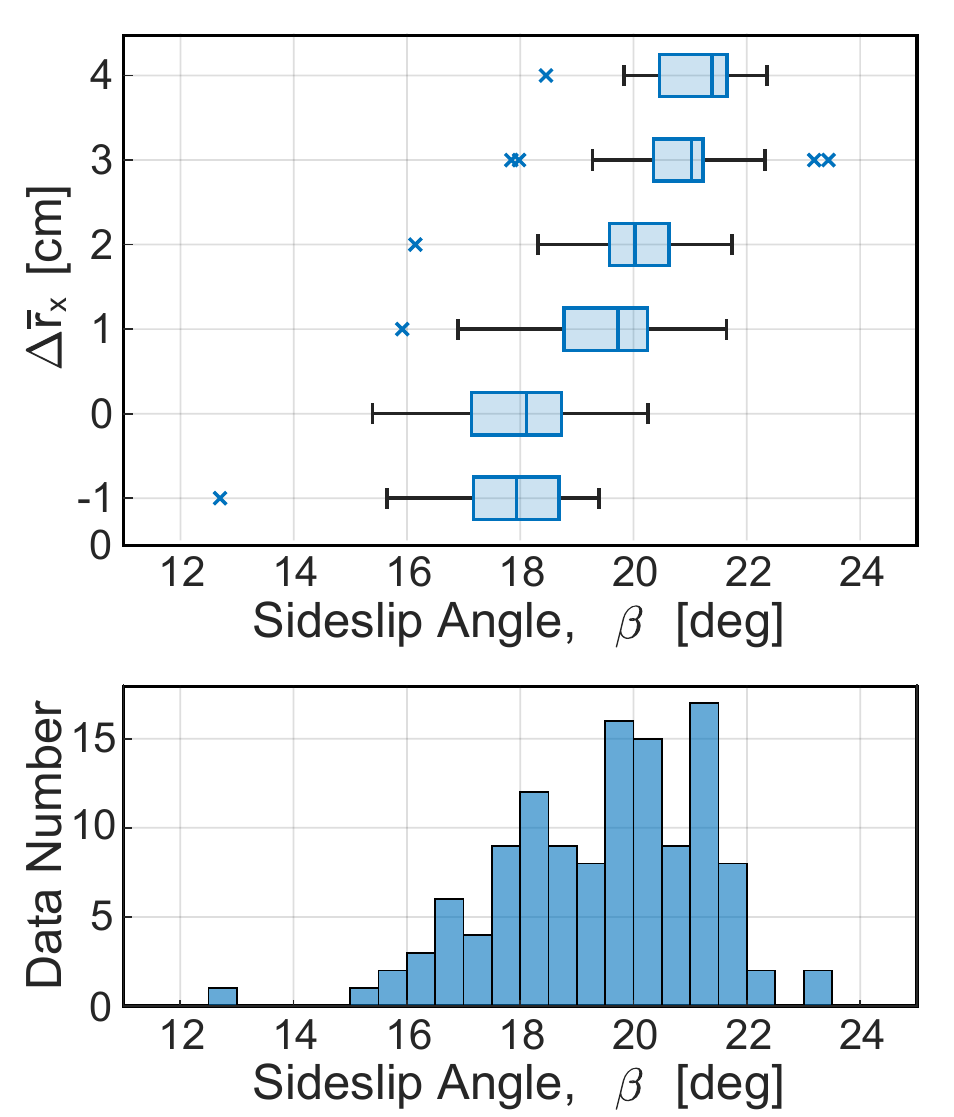}
      }
      \caption{Experimental results of the angle of attack and the sideslip angles in the steady flight of straight-line and spiral motions. 
      (a) In straight-line motion, $\alpha$ changes with $\Delta\bar{r}_x$, and $\beta\approx0$. (b) In spiral motion, $\alpha$ and $\beta$ both change with $\Delta\bar{r}_x$ (and $F_l\!-\!F_r$ as well) $\beta$ is mainly considered. 
      }
      \label{fig.4}
\end{figure}

In the steady straight-line flight, we varied $\Delta\bar{r}_x$ at $\SI{11}{}$ different positions including $-5,...,4,5\,\SI{}{cm}$ while holding the thrust $F_l$ and $F_r$ constant at $\SI{2}{gf}$. 
We repeated the flight experiment $\SI{10}{}$ times for each displacement setting, totaling $\SI{110}{}$ trials. 
In the steady spiral flight, we varied $\Delta\bar{r}_x$ at $\SI{6}{}$ different positions including $-1,...,3,4\,\SI{}{cm}$ and $F_l\!-\!F_r$ at $\SI{6}{}$ differential values including $-3.2,-3.7,-4.2,-4.3,-4.4,-4.9\,\SI{}{gf}$ while holding the $F_l\!+\!F_r$ constant at $\SI{7}{gf}$. 
We repeated the spiraling experiment $\SI{4}{}$ times for each experimental setting. 
Totally $\SI{124}{}$ trials were finally used in system identification with $\SI{20}{}$ excluded as outliers due to the inconsistency in the data, which we conjecture comes from the experimental conditions such as the limited space and the ventilation system. 
All the experiments were recorded and analyzed using the motion capture system. 
Experimental results of the angle of attack and the sideslip angle at the steady state are shown in Fig.\!~\ref{fig.4}. 
\vspace{1mm}

The aerodynamic forces ($D,S,L$) and aerodynamic moments ($M_1,M_2,M_3$) are calculated from the experimental data by solving the set of equations \eqref{eq.26}-\eqref{eq.27} and \eqref{eq.18}-\eqref{eq.19}. 
We identify the mapping from the aerodynamic angles $\alpha$ and $\beta$ to the aerodynamic force and moment coefficients $C_\Box$ and the damping coefficients $K_i$ in \eqref{eq.20}-\eqref{eq.25}. 
Taking the assumption that the spiral motion is symmetric about zero sideslip angle $\beta$, we augment the experimental data on the spiral motion ($\beta>0$) by mirroring all the spiral flight data about the $x\text{-}O\text{-}z$ plane. 
Based on the experimental results and augmented data, we model the aerodynamic force and moment coefficients as polynomial functions of the aerodynamic angles $\alpha$ and $\beta$, and use polynomial regression to identify the model parameters or aerodynamic coefficients, leading to 
\begin{align}
        C_D(\alpha,\!\beta)\ &= C_{D}^0 \hspace{-0.5mm} + {C_D^{\alpha}}\alpha^2 + {C_D^{\beta}}\beta^2\!\!\!,\label{eq.28}\\ 
        C_S(\alpha,\!\beta)\ &= C_{S}^0 + {C_S^{\alpha}}\alpha^2 + {C_S^{\beta}}\beta\,,\\
        C_L(\alpha,\!\beta)\ &= C_{L}^0 + {C_L^{\alpha}\alpha\hspace{1.5mm}+ {C_L^{\beta}}\beta^2}\!\!,\\
	C_{M_1}(\alpha,\!\beta) &= C_{M_1}^0 + {C_{M_1}^{\alpha}}\alpha + {C_{M_1}^{\beta}}\beta\,,\\
	C_{M_2}(\alpha,\!\beta) &= C_{M_2}^0 + {C_{M_2}^{\alpha}}\alpha + {C_{M_2}^{\beta}}\beta^4\!\!,\\
	C_{M_3}(\alpha,\!\beta) &= C_{M_3}^0 +{C_{M_3}^{\alpha}}\alpha + {C_{M_3}^{\beta}}\beta\,.\label{eq.33}
\end{align}
Here, all the constants with ``$C$" in their notations are aero-dynamic coefficients identified from the experimental data. 

\begin{figure}[thpb]
      \centering
      \includegraphics[scale=0.32]{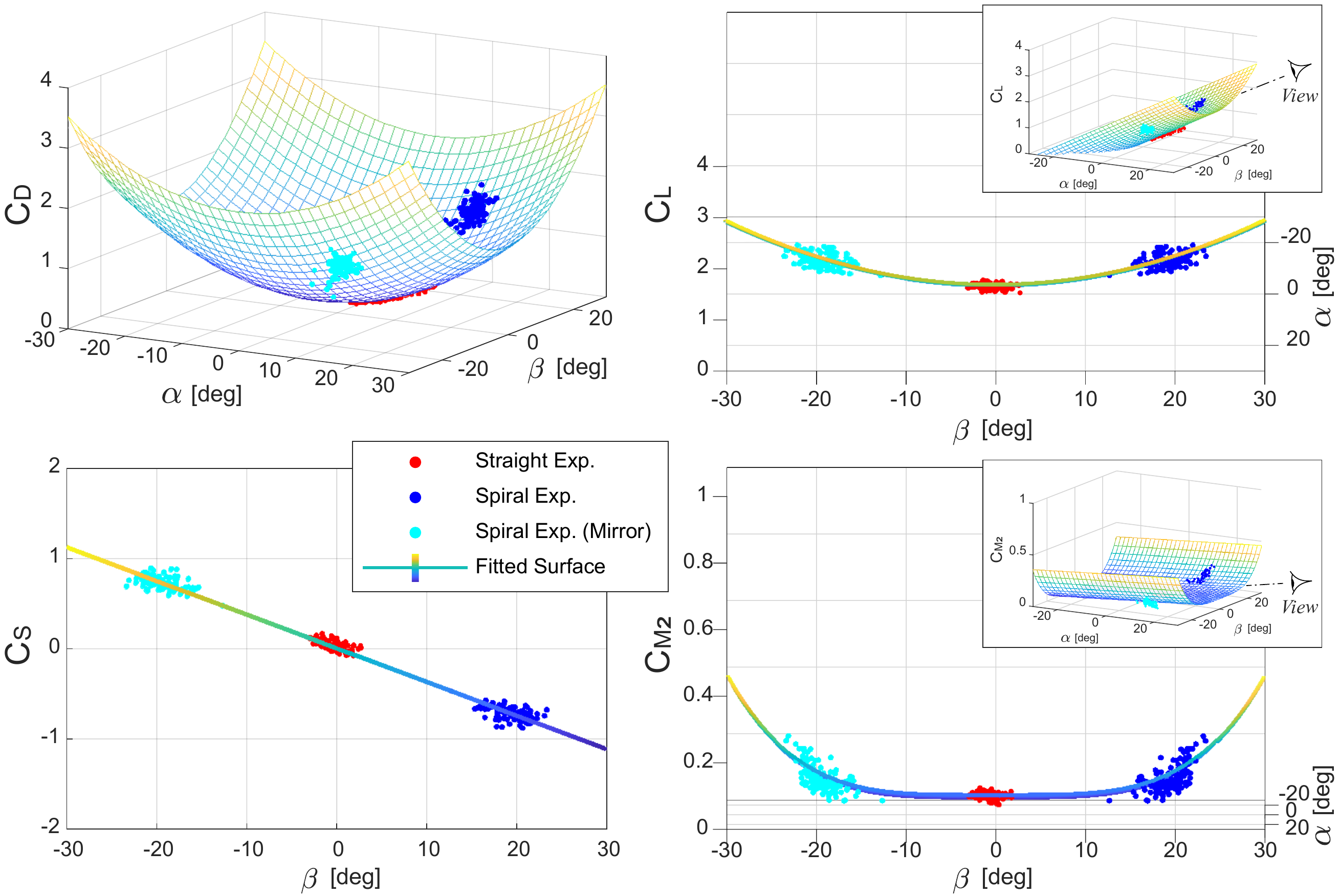}
      \vspace{-0mm}
      \caption{The system identification results of the aerodynamic force and moment coefficients modeled as polynomial functions of the aerodynamic angles $\alpha$ and $\beta$. 
      The drag coefficient $C_D$ is quadratic in both aerodynamic angles. 
      The lift coefficient $C_L$ and the pitching coefficient $C_{M_2}$ are linear in $\alpha$ as in conventional aerial vehicles, but quadratic and quadruplicate with $\beta$ respectively. 
      The side force coefficient $C_S$ is linear in $\beta$ but independent of $\alpha$, which characteristic is also observed in the rolling coefficient $C_{M_1}$ and the yawing coefficient $C_{M_3}$. 
               }
      \label{fig.5}
\end{figure}
\vspace{-0mm}

Solving \eqref{eq.20}-\eqref{eq.25} and \eqref{eq.28}-\eqref{eq.33}, the aerodynamic and damping coefficients are determined using the steady flight experimental and augmented data, with the reference area $A\!=\!(V_\mathrm{He})^{2/3}\!=\!\SI{0.25}{m^2}$. 
We used the nonlinear least squares optimization method, specifically the trust-region-reflective algorithm \cite{algorithemTR}, to fine-tune the coefficients, the results of which are shown in Table\!~\ref{tab:params}.  
\begin{table}[ht]
    \caption{\vspace{-0mm}Identified aerodynamic coefficients of the RGBlimp prototype. }
    \label{tab:params}
    \vspace{-2mm}
    \begin{center}
        \begin{tabular}{p{1.1cm}<{\centering}p{0.6cm}<{\raggedleft}p{2cm}<{\raggedleft}p{2cm}<{\raggedleft}}
            \toprule
            $C_{\hspace{-0.3mm}\Box}^\Box$ & \makecell[c]{$0$} & \makecell[c]{\ \ \ $\alpha$} & \makecell[c]{\ \ \ $\beta$} \\
            \midrule
            $D$   & \SI{0.243}{} & \SI{4.419}{rad^{-2}} & \SI{7.508}{rad^{-2}}\\
            $S$   & \SI{0.001}{} & \SI{-0.074}{rad^{-2}} & \SI{-2.113}{rad^{-1}}\\
            $L$   & \SI{0.159}{} & \SI{2.938}{rad^{-1}} & \SI{4.554}{rad^{-2}}\\
            $M_1$ & \SI{0.001}{} & \SI{-0.030}{rad^{-1}} & \SI{-0.526}{rad^{-1}}\\
            $M_2$ & \SI{0.057}{} & \SI{0.093}{rad^{-1}} & \SI{5.236}{rad^{-4}}\\
            $M_3$ & \SI{0.001}{} & \SI{-0.001}{rad^{-1}} & \SI{-0.093}{rad^{-1}}\\
            \midrule
            \multicolumn{4}{l}{$K_1,K_2,K_3$\hspace{0.5cm}
            \SI{-0.050}{},\hspace{0.1cm}
            \SI{-0.026}{},\hspace{0.1cm}
            \SI{-0.014}{\ \ N\!\cdot\!m\!\cdot\!s/rad}}\\
            \bottomrule
        \end{tabular}
    \end{center}
\end{table}

We want to point out that the adopted aerodynamic model with aerodynamic coefficients is only valid within a certain range of the aerodynamic angles $\alpha$ and $\beta$. 
For example, an excessive angle of attack $\alpha$ (e.g., larger than $\SI{16}{deg}$) typically leads to a flight stall, making the aerodynamic model ineffective. 

\section{Experimental Validation and Analysis}
\label{sec:Experiment}

\subsection{Model Validation}
\label{subsec:Validation}
Extensive experiments of steady and dynamic flights were conducted to validate the established dynamics model and system identification of the RGBlimp prototype. 

\begin{figure}[thpb]
      \centering
      \subfigbottomskip=1pt
      \subfigure[Pitch Angle $\theta$ vs. the moving mass displacement $\Delta\bar{r}_x$.]{
      \label{fig.6a}
      \includegraphics[scale=0.52]{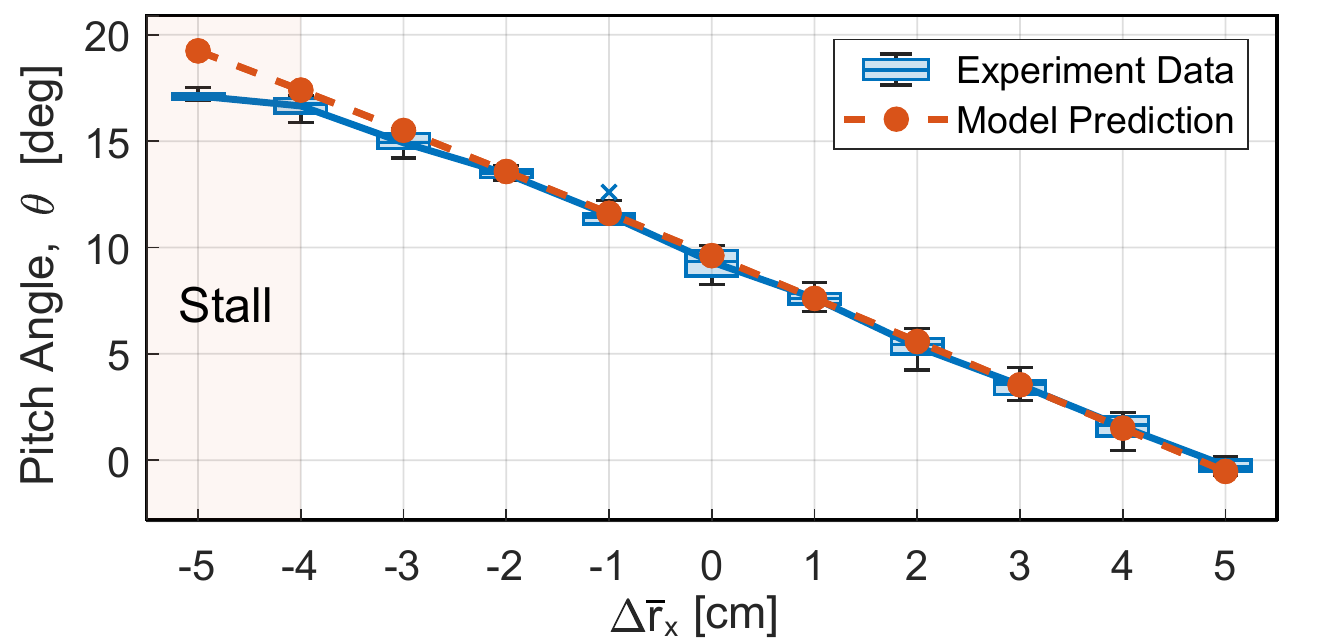}
      }
      \subfigure[Flight speed $V$ vs. the moving mass displacement $\Delta\bar{r}_x$.]{
      \label{fig.6b}
      \includegraphics[scale=0.52]{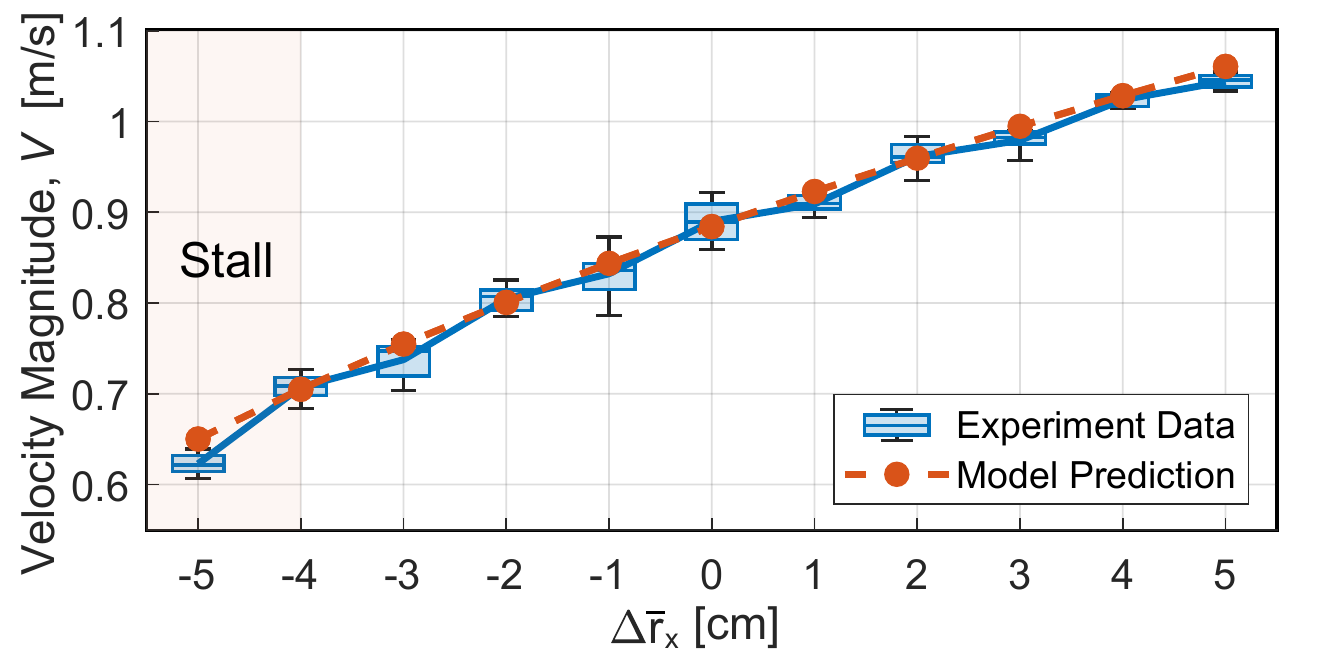}
      }
      \caption{Experimental results of the steady straight-line flight with a good match between the model prediction and the experimental data. 
      The box charts show the medians, maxima, minima, and quartiles of $\theta$ and $V$ over $\SI{10}{}$ tests given a fixed $\Delta\bar{r}_x$. } 
      \label{fig.6}
\end{figure}

\textit{1)  Steady flight}: 
first, we used the planar straight-line flight test, characterized by two variables including the pitching angle $\theta$ and the velocity magnitude $V$. 
We varied the moving mass displacement $\Delta\bar{r}_x$ and kept both thrusts $F_l$ and $F_r$ the same to obtain various steady flight trajectories. 
Figure\!~\ref{fig.6} shows the comparison results between the model prediction and the experimental data. 
\vspace{-0mm}

We observe that $\theta$ and $V$ from the model prediction match the experimental data reasonably well. 
In addition, there exists a significant mismatch between the model prediction and the experimental results when the moving mass moves far back, resulting in a flight stall. 
\begin{figure}[thpb]
      \centering
      \includegraphics[scale=0.52]{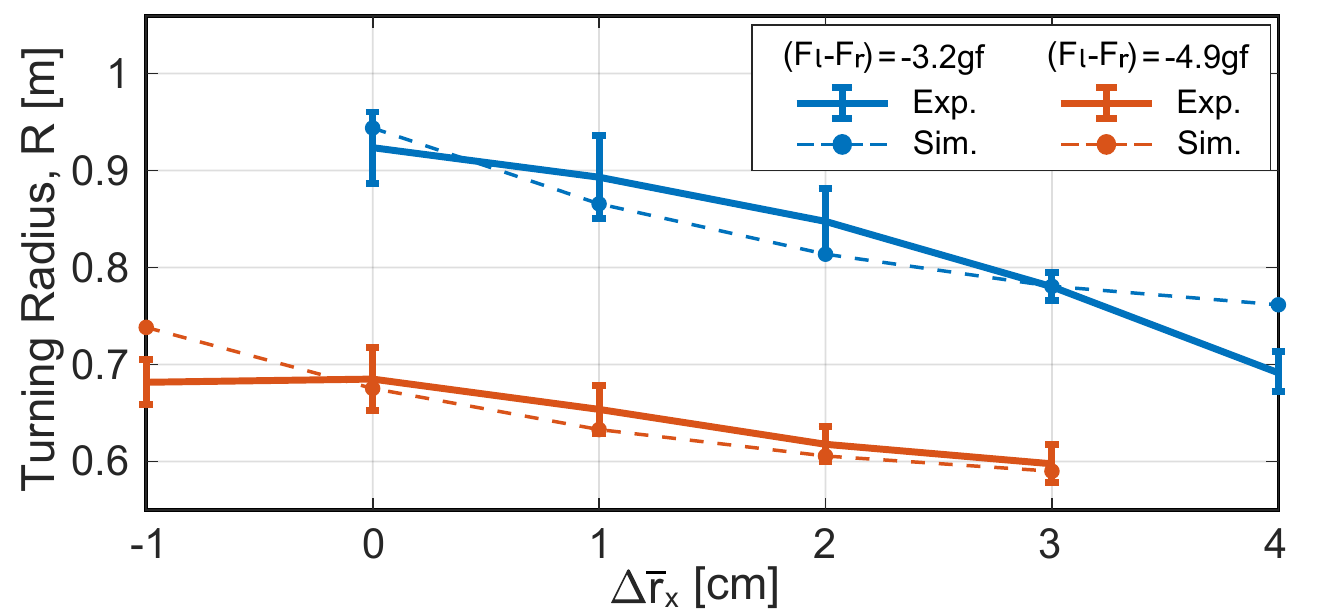}
      \vspace{-0mm}
      \caption{Experimental results of the steady spiraling flight with a good match between the model prediction and the experimental data. The error bar shows the average and standard deviation of radius $R$ over $\SI{4}{}$ tests given a fixed $\Delta\bar{r}_x$. 
       }
      \label{fig.7}
\end{figure}
\vspace{-0mm}

We also conducted steady spiral flight tests where we focused on the radius of curvature of the projected flight trajectory onto the horizontal plane, denoted by $R$. 
The experimental results with different moving mass displacements $\Delta\bar{r}_x$ and propulsion thrusts $F_l$ and $F_r$ are shown in Fig.\!~\ref{fig.7}. 
The error bar shows the averaged value and standard deviation of the turning radius of curvature $R$ over $\SI{4}{}$ independent trials. 
With modeling uncertainties and environmental disturbances, we consider the match between the model predictions and experimental results satisfactory. 

\begin{figure}[thpb]
      \centering
      \subfigbottomskip=0.1pt
      \subfigure[Flight trajectory. Left: top view, right: 3D view.]{
      \label{fig.8a}
      \includegraphics[scale=0.36]{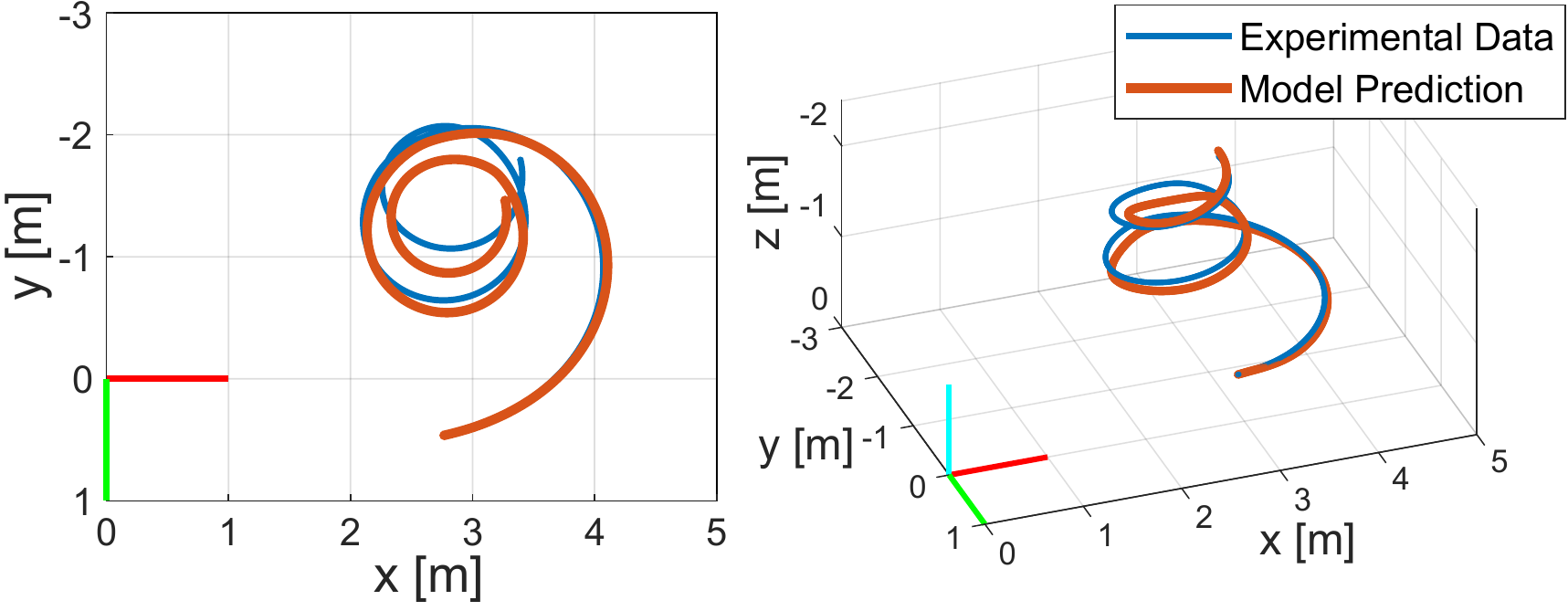} 
      } \hspace{-5mm}
      \subfigure[Thrust inputs.]{
      \label{fig.8b}
      \includegraphics[scale=0.36]{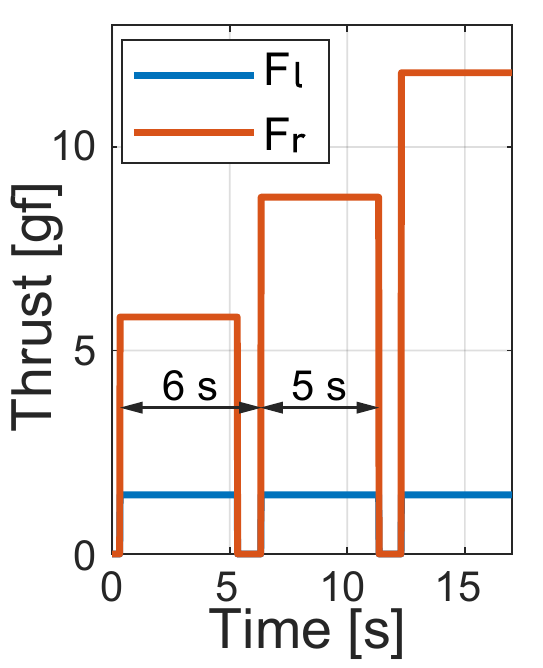}
      }
      \hspace{-5mm}
      \subfigure[Turning radius $R$ and vertical speed $V_z$.]{
      \label{fig.8c}
      \includegraphics[scale=0.36]{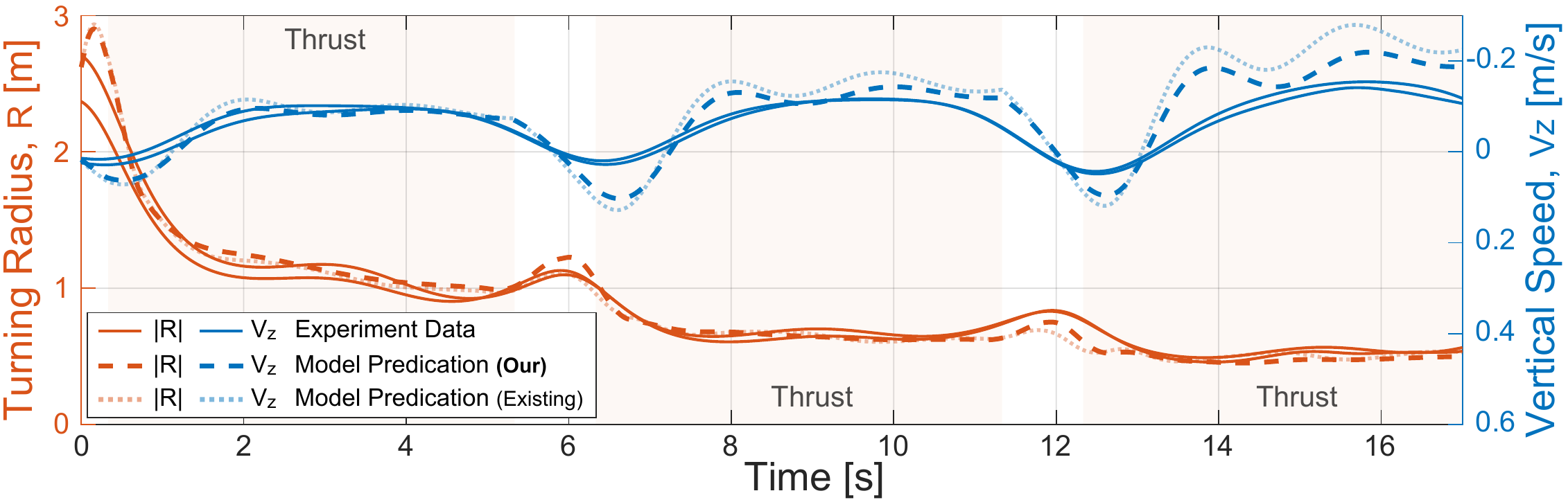}
      }
      \caption{Experimental results of the dynamic flight under a time-varying thrust input along with model prediction. 
              (a) The flight trajectories in top view and 3D view. 
              (b) The trajectories of the thrust control inputs $F_l$ and $F_r$, while holding $\Delta\bar{r}_x=\SI{-1}{cm}$. 
              (c) The comparison in the turning radius $R$ and vertical speed $V_z$ between the experimental results of two independent tests, the simulation results of the proposed model in this paper, and the commonly-used model in the literature that does not consider the non-zero displacement of the CG from the CB. }
      \label{fig.8}
\end{figure}

\textit{2) Dynamic flight}: 
spiral flight experiments with time-varying control inputs were conducted to validate the established model. 
We used a staircase function for the right thrust $F_r$ while keeping the left thrust $F_l$ a constant, expecting to generate spiral trajectories with time-varying turning radius $R$. 
The experimental results along with the model prediction are presented in Fig.\!~\ref{fig.8}. 

From the experimental results, we observe the proposed model predicts well the dynamic motion of the RGBlimp in both the powered phase and the unpowered gliding phase. 
We want to point out that there exists a mismatch in the oscillatory motion especially at higher flight speed. We conjecture that this mismatch comes from the modeling simplification of the nonlinear variation of the rotational damping with the flight speed. 
Moreover, the commonly-used model for miniature blimps \cite{GTMAB2} is simulated as a comparison, which ignores the Coriolis force and centrifugal force that are caused by the separation between the CG and CB, leading to a decreased modeling accuracy, compared to our established dynamic model. 

\subsection{Aerodynamic Analysis}
\label{subsec:aeroAna}
The design of the RGBlimp utilizes both a lighter-than-air robot body and a pair of fixed wings to generate the lift, which is expected to benefit not only the aerodynamic efficiency but also the aerodynamic stability of the robot. 
This section conducts an analysis of the RGBlimp aerodynamics based on the RGBlimp prototype and a comparison counterpart prototype of the same design with no wings. 

\begin{figure}[thpb]
      \centering
      \vspace{-0mm}
      \subfigbottomskip=0pt
      \subfigure[Gliding trajectories in the sagittal plane when $\Delta\bar{r}_x=\SI{5}{cm}$. ]{ 
      \label{fig.p1a}
      \includegraphics[scale=0.28]{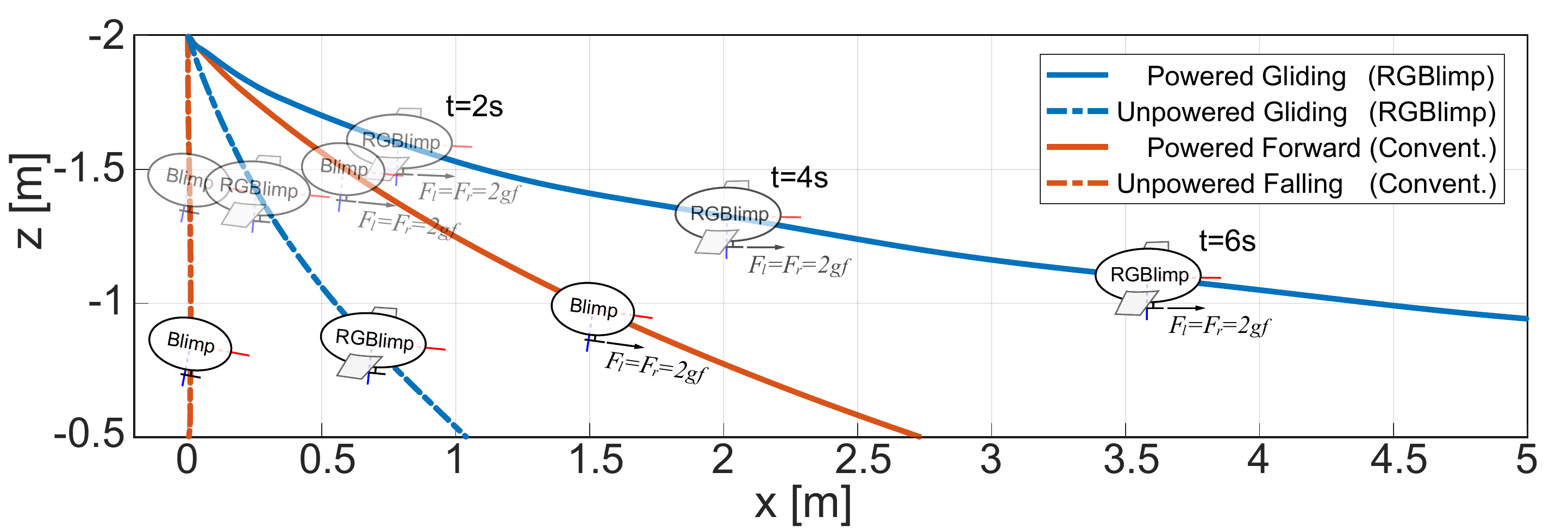}\hspace{3mm} 
      } 
      \subfigure[L/D ratio in static flight vs. the moving mass displacement $\Delta\bar{r}_x$. ]{ 
      \label{fig.p1b}
      \includegraphics[scale=0.28]{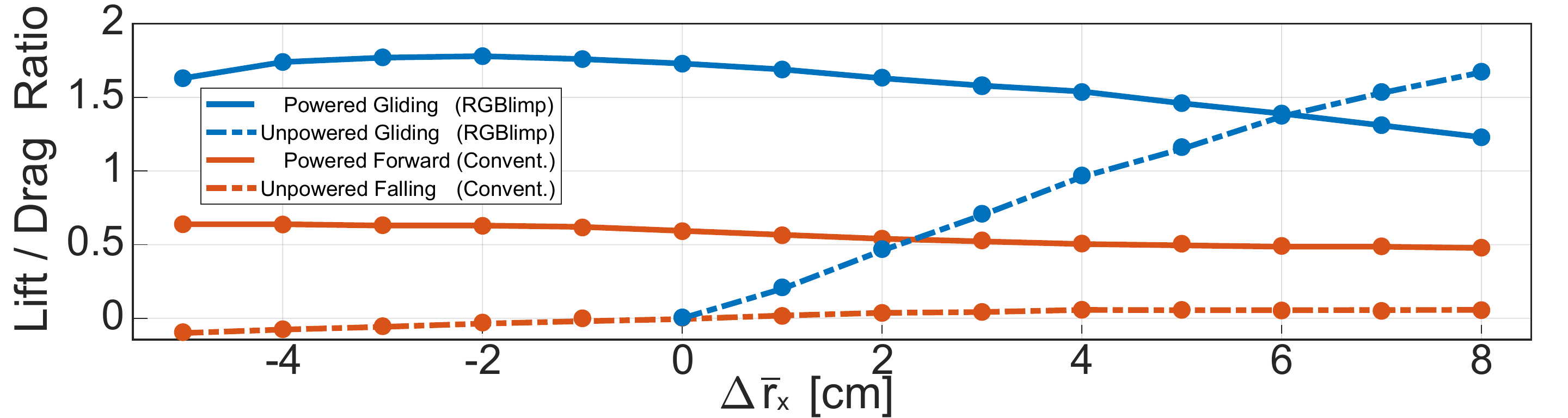}\hspace{3mm} 
      }
      \caption{The comparison results of the gliding performance in experiment between the RGBlimp and its counterpart following the conventional design of no wings in powered ($F_l\!=\!F_r\!=\!\SI{2}{gf}$) and unpowered gliding. 
      }
      \label{fig.p1}
\end{figure}

\textit{1) Gliding performance}: 
an aerial robot is called gliding when it takes advantage of aerodynamics for its forward motion and maintains a relative stable descending flight. 
Depending on whether the robot uses propulsion power or not, we have powered gliding and unpowered gliding, respectively. 
The gliding performance is typically measured by the relationship between forward speed and descending speed. 
Figure~\ref{fig.p1} shows the experimental trajectories of both the RGBlimp and the comparison counterpart in powered and unpowered gliding, and their L/D ratio in static flight with different moving mass displacements $\Delta\bar{r}_x$. 
It is observed that the fixed wings of the RGBlimp prototype significantly improve the gliding performance compared to the conventional robotic blimp with no wings. 

\begin{figure}[thpb]
      \centering
      \vspace{-1mm}
      \subfigbottomskip=2pt
      \subfigure[$C_L$ vs. $\alpha$.]{
      \label{fig.9a}
      \includegraphics[scale=0.41]{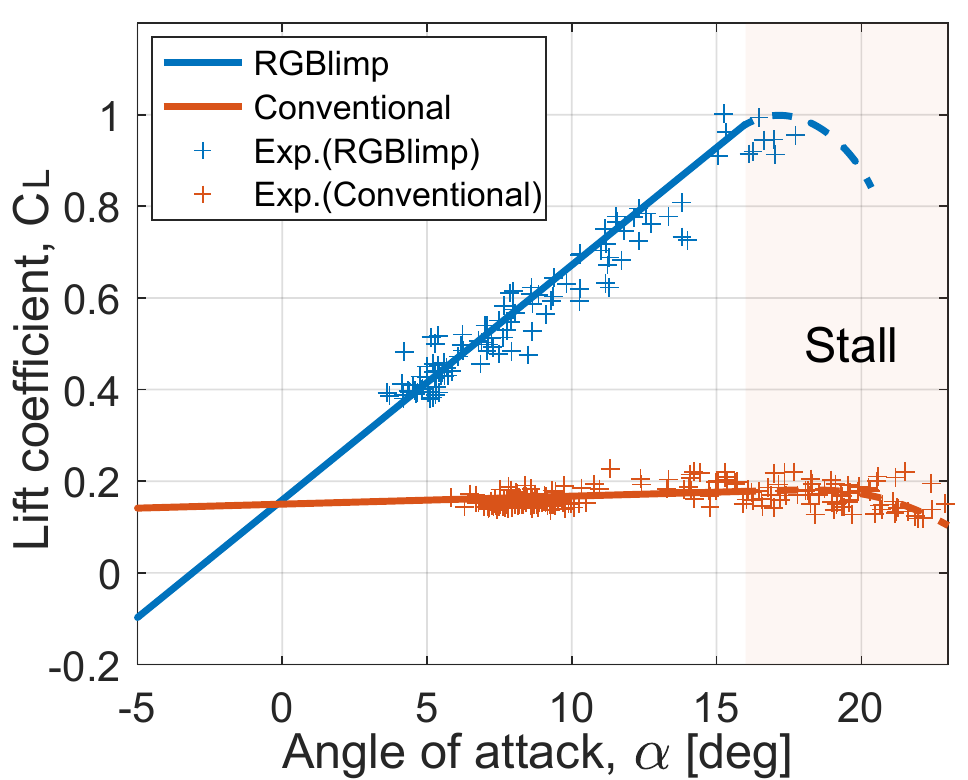} 
      } \hspace{-2mm}
      \subfigure[$C_D$ vs. $\alpha$.]{
      \label{fig.9b}
      \includegraphics[scale=0.41]{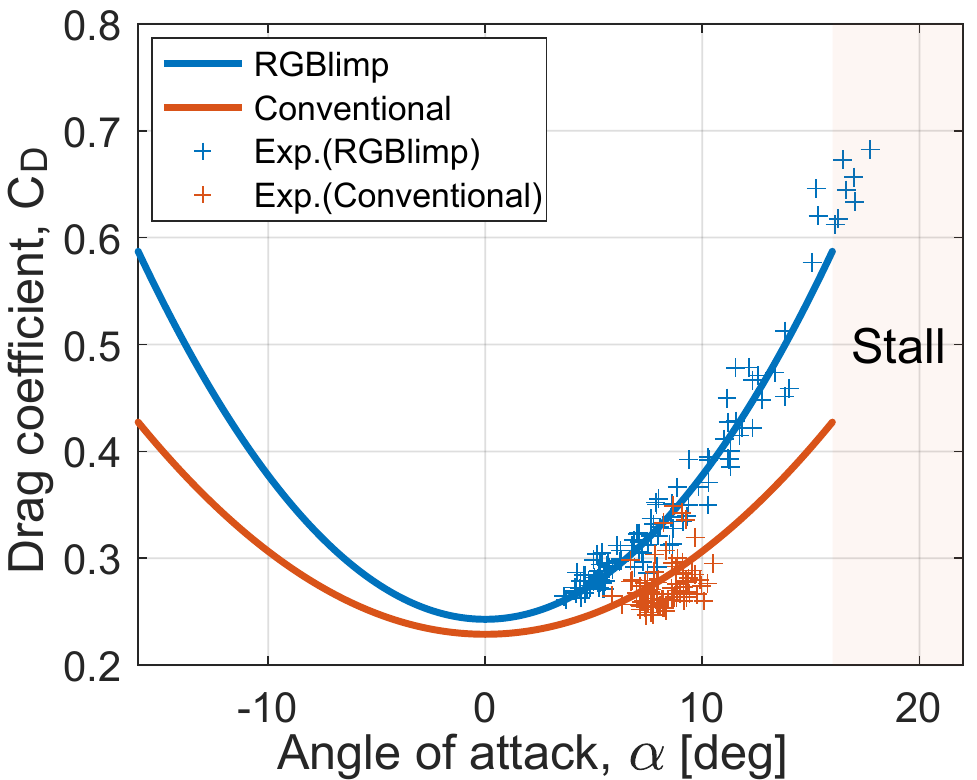}
      }
      \caption{The comparison results of the lift and drag coefficients between the RGBlimp and its counterpart following a conventional design with no wings. 
      (a) The fixed wings significantly improve the lift coefficient in the RGBlimp when compared to conventional blimps. 
      (b) The drag coefficient of the RGBlimp is slightly greater than that of conventional blimps for the additional drag from the wings. 
      Note that the dotted lines indicate possible curves in the stall. 
      }
      \label{fig.9}
\end{figure}

\textit{2) Aerodynamic efficiency}: 
the coefficients of aerodynamic lift and drag $C_L$ and $C_D$ are used as indicators to investigate the aerodynamic efficiency of the robot. 
Figure\!~\ref{fig.9} shows the experimental results and the model prediction of $C_L$ and $C_D$ as functions of the angle of attack $\alpha$, when tested at a flight speed of $\SI{1}{m/s}$. 
It is observed that the fixed wings of the prototype lead to a significant increase in the aerodynamic lift and a slight increase in the aerodynamic drag when compared to the conventional robotic blimp that has no wings. 
The aerodynamic lift drops sharply after the angle of attack is greater than $\SI{16}{deg}$, which corresponds to the stall phenomenon when the moving mass is positioned far back shown in Fig.\!~\ref{fig.6}. 

\begin{figure}[thpb]
      \centering
      \vspace{-2mm}
      \subfigbottomskip=2pt
      \subfigure[L/D ratio. ]{
      \label{fig.10a}
      \includegraphics[scale=0.370]{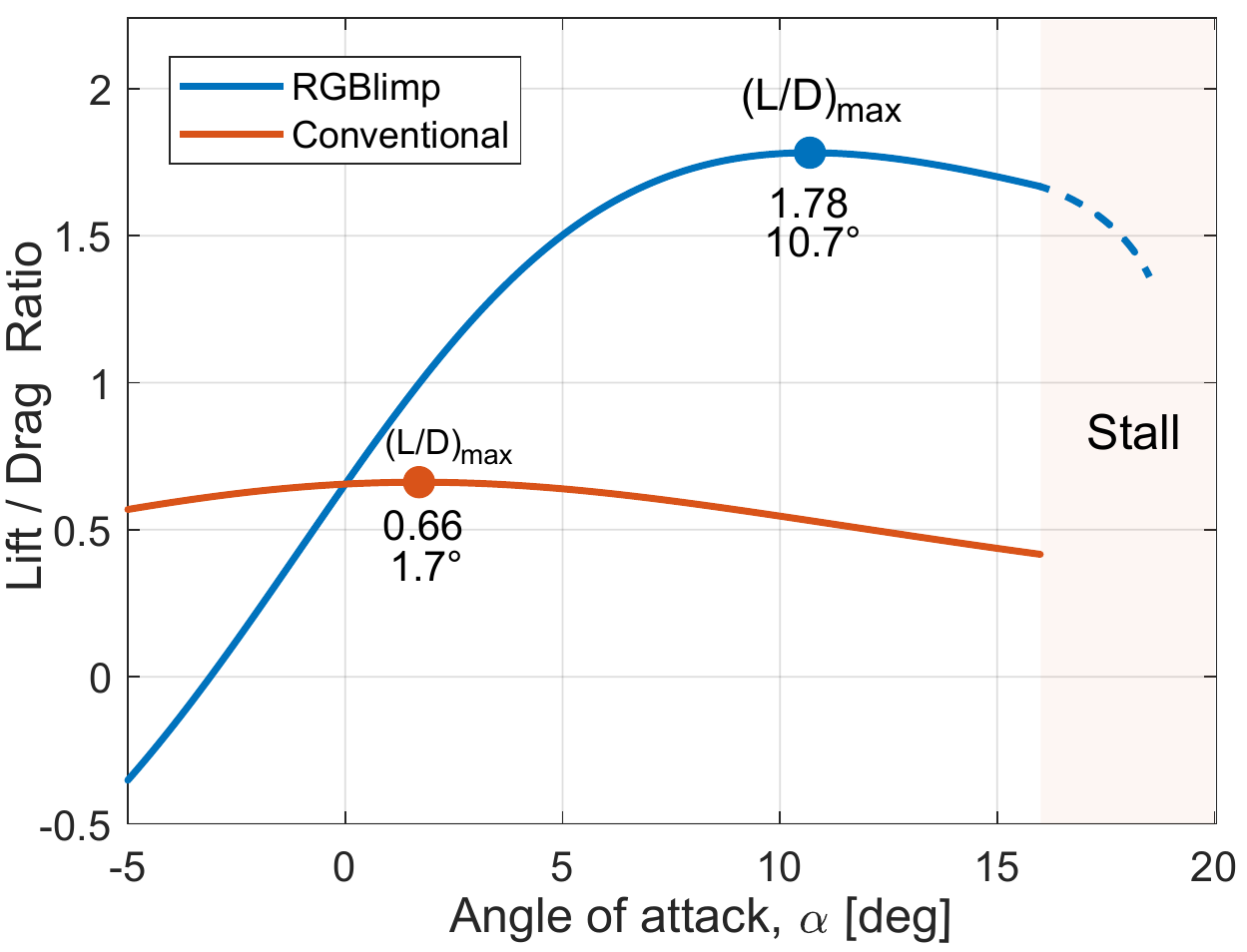} 
      } \hspace{-5mm}
      \subfigure[Drag polar.]{
      \label{fig.10b}
      \includegraphics[scale=0.370]{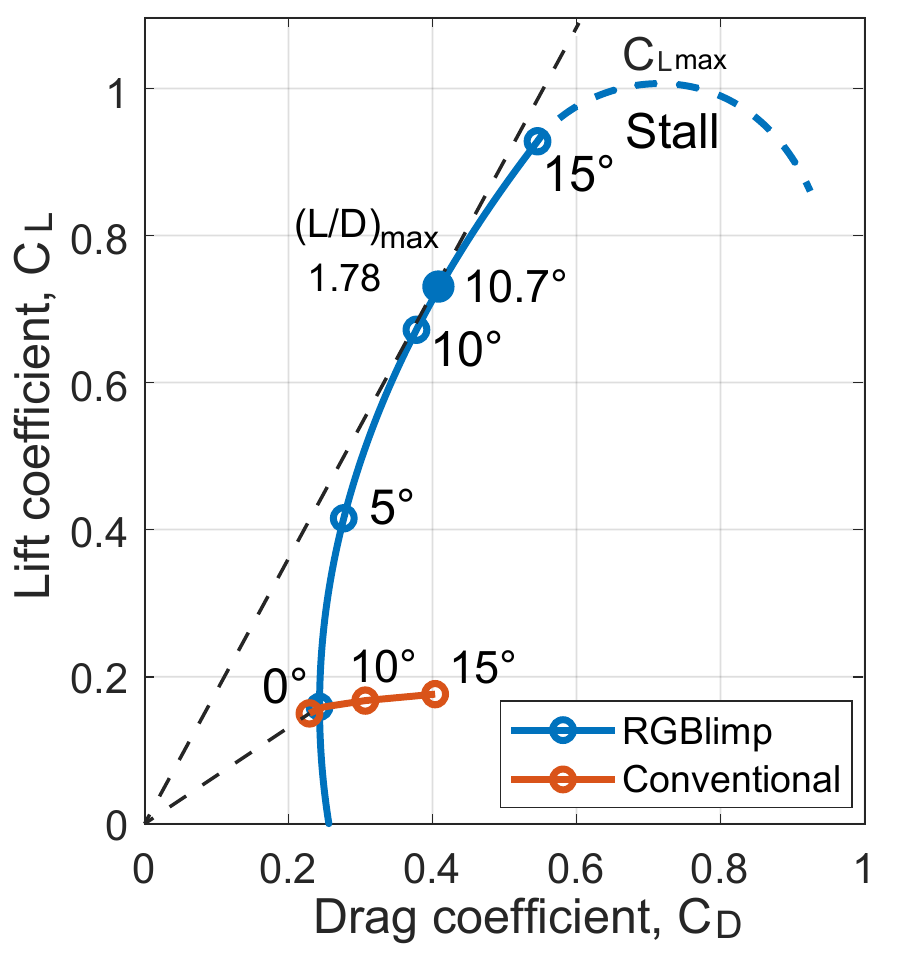}
      }
      \caption{The comparison results of the L/D ratio and the drag polar between the RGBlimp and its counterpart following a conventional design with no wings. 
      The maximum L/D ratio of the RGBlimp is $\SI{1.78}{}$ when the angle of attack $\alpha$ is $\SI{10.7}{deg}$, which is twice higher than conventional blimps. 
      Note that the dotted lines indicate possible curves in the stall. 
      }
      \label{fig.10}
\end{figure}

To evaluate the combined effects of both lift and drag on aerodynamic efficiency, the L/D ratio and the drag polar are plotted in Fig.\!~\ref{fig.10}. 
It is observed that the L/D ratio of the RGBlimp reaches a maximum value of $\SI{1.78}{}$ when the angle of attack $\alpha$ is $\SI{10.7}{deg}$, while that of the counterpart design with no wings is only about $1/3$ at $0.66$. 
The drag polar via the tangent line to the curve through the origin demonstrates that there exists an increased lift in the RGBlimp when compared to its counterpart at every drag coefficient. 
The maximum L/D ratio stands for optimal flight efficiency. 

The aerodynamic lift $L_\mathrm{aero}$ is a solid supplement to the buoyant lift $L_\mathrm{buoyant}$ of the RGBlimp. 
Take the flight with the maximum L/D ratio of $\SI{1.78}{}$ and a cruising speed at $\SI{1}{m/s}$ as an example. 
The total lift is calculated as the sum of $L_\mathrm{aero}$ and $L_\mathrm{buoyant}$, i.e., 
\begin{align}
      \displaystyle
      L_\mathrm{total} = L_\mathrm{aero} + L_\mathrm{buoyant} = \SI{11}{gf} + \SI{152}{gf} = \SI{163}{gf}. \nonumber
\end{align}
The aerodynamic lift takes $\SI{6.7}{\%}$ of the total lift. 

\begin{figure}[thpb]
      \centering
      \vspace{-0mm}
      \subfigbottomskip=2pt
      \subfigure[$C_{M_2}$ vs. $\alpha$.]{
      \label{fig.11a}
      \includegraphics[scale=0.380]{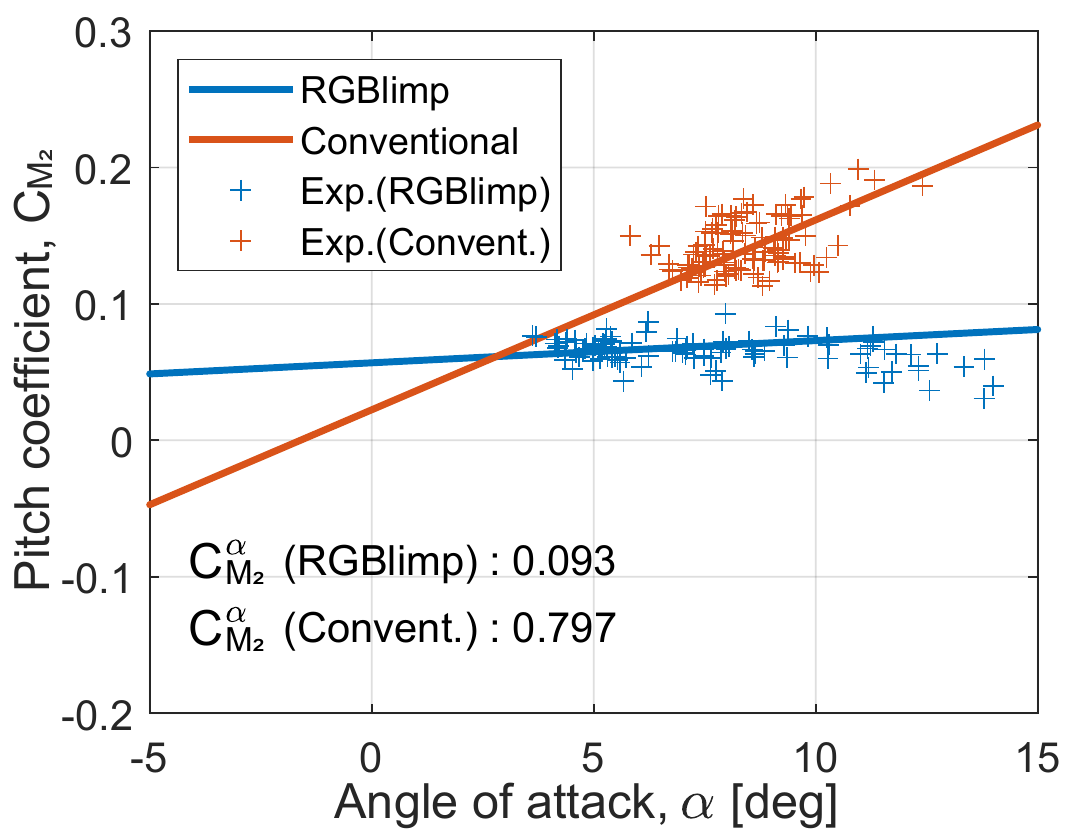} 
      } \hspace{-5mm}
      \subfigure[$C_{M_3}$ vs. $\beta$.]{
      \label{fig.11b}
      \includegraphics[scale=0.380]{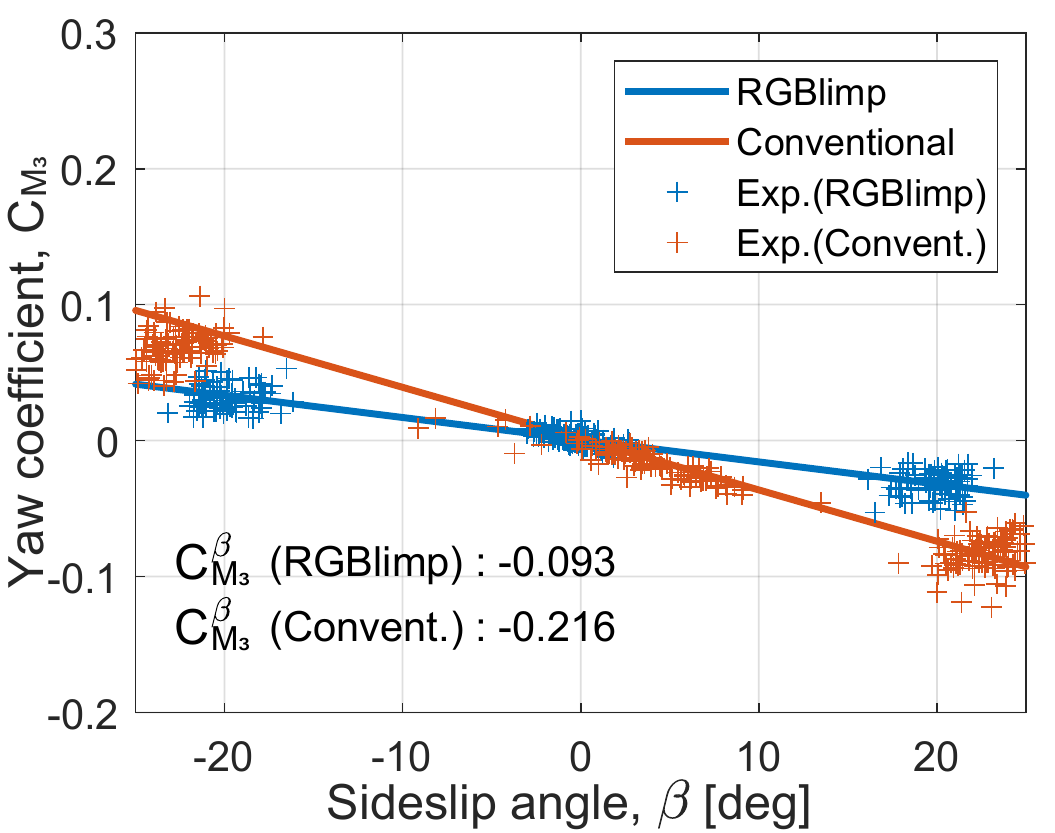}
      }
      \caption{The comparison results of the pitch and yaw moment coefficients between the RGBlimp prototype and its counterpart following a conventional design with no wings. 
      (a) The main wings effectively reduce the rise of pitch moment with the angle of attack, thus improving the longitudinal stability concerning pitch motion. 
      (b) The yaw moment shows a similar effect, improving the directional stability concerning yaw motion. 
      }
      \label{fig.11}
\end{figure}

\textit{3) Aerodynamic stability}: 
we consider the static stability in the pitch and yaw motion, also known as the longitudinal and directional stability of aerial robots. 
The static stability reflects the ability of an aircraft recovering from an impulse disturbance. 
The pitch and yaw moment coefficients $C_{M_2}^\alpha$ and $C_{M_3}^\beta$ are used to measure the static stability in pitch and yaw motion, respectively, as shown in Fig.\!~\ref{fig.11}. 
\vspace{-0mm}

Both the RGBlimp prototype and its counterpart design with no wings possess a positive $C_{M_2}^\alpha$ and a negative $C_{M_3}^\beta$, which have a negative effect on the longitudinal and directional stability, respectively. 
The addition of the fixed wings leads to a more gentle slope in the $C_{M_2}^\alpha$ and $C_{M_3}^\beta$ graphs as shown in Fig.\!~\ref{fig.11}, thus improving the corresponding flight stability. 
For both the steady straight-line and spiral flights, we investigate the local stability of the dynamic system by checking the linearized system about the eigenvalue distribution. 
While the system is locally asymptotically stable at each of the steady flight equilibrium points with the linearization matrix $\boldsymbol{A}$ Hurwitz, the RGBlimp system, compared to the counterpart system with no wings, has a larger magnitude in the negative real parts of the eigenvalues of matrix $\boldsymbol{A}$, corresponding to a faster convergence speed towards the system equilibrium or the steady flight. 
For example, consider the straight-line flight motion with inputs $F_l\!=\!F_r\!=\!\SI{2}{gf}$ and $\Delta\bar{r}_x\!=\!\SI{0}{cm}$, the slowest mode of the RGBlimp dynamic system has an eigenvalue of $\SI{-0.37}{s^{-1}}$, providing a much faster convergence compared to the system with no wings that has an eigenvalue of $\SI{-0.06}{s^{-1}}$. 
As a result, the undesired long-standing swinging motion of a conventional robotic blimp due to the poor longitudinal stability, as reported in \cite{systemId}\cite{swing}, is not observed as a concern in the RGBlimp. 
Here we want to point out that the aerodynamic stability is also affected by $\Delta r_x$, which could potentially be used for flight control.

\section{Conclusions}
\label{sec:Conclusion}

This letter presented a novel design and derived the dynamic model of a miniature robotic blimp named RGBlimp. 
Aiming to combine the strengths of both the aerodynamic and buoyant lift, the RGBlimp combined a pair of fixed wings and an LTA body for lift generation, and a moving mass and a pair of propellers for actuation control. 
An RGBlimp prototype was developed and system identification was conducted using nonlinear regression over steady flight test data. 
Extensive flight experiments were conducted to validate the proposed design and the derived model. 
Comparative analysis between the aerodynamics of the RGBlimp prototype and that of a counterpart design with no wings was presented, revealing a considerable performance improvement of the RGBlimp in aerodynamic efficiency and aerodynamic stability. 

In future work, we plan to instrument the RGBlimp with some inertial measurement units and a vision sensor, and investigate the closed-loop control in low-speed cruising. 
We will take the wind disturbance into account when designing the controller for a robust performance in real-world applications such as outdoor environmental monitoring.

\section*{Acknowledgment}
The authors would like to thank Prof. Zhongkui Li for his help in the motion capture experiment.

\ifCLASSOPTIONcaptionsoff
  \newpage
\fi




\begin{thebibliography}{99}
\bibitem{GTMAB2} S. Cho et al., ``Autopilot design for a class of miniature autonomous blimps," in \textit{IEEE Conf. on Control Technology and Applications (CCTA)}, pp.841-846, 2017, doi: \href{https://doi.org/10.1109/CCTA.2017.8062564}{10.1109/CCTA.2017.8062564}. 
\bibitem{modelBlimpICRA98} S. B. V. Gomes and J. G. Ramos, ``Airship dynamic modeling for autonomous operation," in \textit{IEEE International Conference on Robotics and Automation (ICRA)}, vol. 4, pp. 3462-3467, 1998, doi:\href{https://doi.org/10.1109/ROBOT.1998.680973}{10.1109/ROBOT.1998.680973}. 
\bibitem{BlimpIROS00} S. Zwann et al., ``Vision based station keeping and docking for an aerial blimp," in \textit{IEEE/RSJ International Conference on Intelligent Robots and Systems (IROS)}, vol. 1, pp. 614-619, 2000, doi: \href{https://doi.org/10.1109/IROS.2000.894672}{10.1109/IROS.2000.894672}. 
\bibitem{GTMAB} Q. Tao et al., ``Modeling and identification of coupled translational and rotational motion of underactuated indoor miniature autonomous blimps," in \textit{International Conference on Control, Automation, Robotics and Vision (ICARCV)}, pp. 339-344, 2020, doi: \href{https://doi.org/10.1109/ICARCV50220.2020.9305371}{10.1109/ICARCV50220.2020.9305371}. 
\bibitem{blimpICRA19} Y. Wang et al., ``Disturbance compensation based control for an indoor blimp robot," in \textit{IEEE International Conference on Robotics and Automation (ICRA)}, pp. 2040-2046, 2019, doi: \href{https://doi.org/10.1109/ICRA.2019.8793535}{10.1109/ICRA.2019.8793535}. 
\bibitem{darpa} C. Lu et al., ``A Heterogeneous Unmanned Ground Vehicle and Blimp Robot Team for Search and Rescue using Data-driven Autonomy and Communication-aware Navigation," in \textit{Field Robotics}, vol. 2, pp. 557-594, 2022, doi: \href{https://doi.org/10.55417/fr.2022020}{10.55417/fr.2022020}. 
\bibitem{blimpMonitoring} S. Badia et al., ``A Biologically Based Flight Control System for a Blimp-based UAV," in \textit{IEEE International Conference on Robotics and Automation (ICRA)}, pp. 3053-3059, 2005, doi: \href{https://doi.org/10.1109/ROBOT.2005.1570579}{10.1109/ROBOT.2005.1570579}. 
\bibitem{blimpIROS13} M. Burri et al., ``Design and control of a spherical omnidirectional blimp," in \textit{IEEE/RSJ International Conference on Intelligent Robots and Systems (IROS)}, pp. 1873-1879, 2013, doi: \href{https://doi.org/10.1109/IROS.2013.6696604}{10.1109/IROS.2013.6696604}. 
\bibitem{blimpMuseum} D. St-Onge et al., ``Dynamic modelling and control of a cubic flying blimp using external motion capture," in \textit{Journal of Systems and Control Engineering}, vol. 229, 2015, doi: \href{https://doi.org/10.1177/0959651815597603}{10.1177/0959651815597603}. 
\bibitem{blimpProjector} J. Lee et al., ``Stabilization of floor projection image with soft unmanned aerial vehicle projector," in \textit{Int. Conf. on Ubiquitous Information Management and Communication (IMCOM)}, pp. 1-5, 2021, \href{https://doi.org/10.1109/IMCOM51814.2021.9377365}{10.1109/IMCOM51814.2021.9377365}. 
\bibitem{DJI} ``DJI Phantom 4 Pro." https://www.dji.com/phantom-4-pro-v2. 
\bibitem{fixedwingSurvey} H. Chao et al., ``Autopilots for small fixed-wing unmanned air vehicles: A survey," in \textit{International Conference on Mechatronics and Automation (ICMA)}, pp. 3144-3149, 2007, doi: \href{https://doi.org/10.1109/ICMA.2007.4304064}{10.1109/ICMA.2007.4304064}. 
\bibitem{HA-aero} A. D. Andan et al., ``Aerodynamics of a hybrid airship," in \textit{AIP Conference Proceedings}, vol. 1440, pp. 154-161, 2012, \href{https://doi.org/10.1063/1.4704214}{10.1063/1.4704214}. 
\bibitem{HA-ex} A. D. Andan et al., ``Investigation of Aerodynamic Parameters of a Hybrid Airship," in \textit{Journal of Aircraft}, vol. 49, pp. 658–662, 2012, \href{https://doi.org/10.2514/1.C031491}{10.2514/1.C031491}. 
\bibitem{HybridSurvey} M. Manikandan et al., ``Research and advancements in hybrid airships—A review," in \textit{Progress in Aerospace Sciences}, vol. 127, 2021, doi: \href{https://doi.org/10.1016/j.paerosci.2021.100741}{10.1016/j.paerosci.2021.100741}. 
\bibitem{bio} K. L. Bishop, ``Aerodynamic force generation, performance and control of body orientation during gliding in sugar gliders (Petaurus breviceps)," in \textit{Journal of Experimental Biology}, vol. 15, 2007, doi: \href{https://doi.org/10.1242/jeb.002071}{10.1242/jeb.002071}. 
\bibitem{flightDynamics} R. F. Stengel, \textit{Flight dynamics}, Princeton University Press, 2022. 
\bibitem{mechanics} J. E. Marsden, T. S. Ratiu, \textit{Introduction to Mechanics and Symmetry: A Basic Exposition of Classical Mechanical Systems}, Springer, 1999.
\bibitem{book1993} T. I. Fossen, \textit{Guidance and Control of Ocean Vehicles}, Wiley, 1993. 
\bibitem{glidingfish} F. Zhang et al., ``Tail-Enabled Spiraling Maneuver for Gliding Robotic Fish," in \textit{ASME. J. Dyn. Sys., Meas., Control}, vol. 136, 2014, doi: \href{https://doi.org/10.1115/1.4026965}{10.1115/1.4026965}.
\bibitem{dynamic} W. C. Durham, \textit{Aircraft Flight Dynamics and Control}, Wiley, 2013. 
\bibitem{aircraft} J. D. Anderson, \textit{Aircraft Performance and Design}, McGraw-Hill, 1998. 
\bibitem{aeroID1} M. Ashraf and M. Choudhry, ``Dynamic modeling of the airship with Matlab using geometrical aerodynamic parameters," in \textit{Aerosp. Sci. Technol.}, vol. 25, 2013, \href{https://doi.org/10.1016/j.ast.2011.08.014}{10.1016/j.ast.2011.08.014}. 
\bibitem{aeroID2} J. Shen et al., ``Calculation and Identification of the Aerodynamic Parameters for Small-Scaled Fixed-Wing UAVs," in \textit{Sensors}, 2018, \href{https://doi.org/10.3390/s18010206}{10.3390/s18010206}. 
\bibitem{algorithemTR} Moré, J.J., and D.C. Sorensen, ``Computing a Trust Region Step," in \textit{SIAM Journal on Scientific and Statistical Computing}, vol. 3, pp. 553–572, 1983, \href{https://doi.org/10.1137/0904038}{10.1137/0904038}. 
\bibitem{systemId} Q. Tao et al., ``Parameter Identification of Blimp Dynamics through Swinging Motion," in \textit{Int. Conf. on Control, Automation, Robotics and Vision (ICARCV)}, pp. 1186-1191, 2018, \href{https://doi.org/10.1109/ICARCV.2018.8581376}{10.1109/ICARCV.2018.8581376}. 
\bibitem{swing} Q. Tao et al., ``Swing-Reducing Flight Control System for an Underactuated Indoor Miniature Autonomous Blimp," in \textit{IEEE/ASME Transactions on Mechatronics}, pp. 1895-1904, 2021, \href{https://doi.org/10.1109/TMECH.2021.3073966}{10.1109/TMECH.2021.3073966}. 
 

\end{thebibliography}
%

\end{document}